\documentclass[10pt,twocolumn,letterpaper]{article}

\usepackage[pagenumbers]{cvpr} % To force page numbers, e.g. for an arXiv version
\usepackage[accsupp]{axessibility}

\usepackage{graphicx}
\usepackage{amsmath}
\usepackage{amssymb}
\usepackage{color}
\usepackage{multirow}
\usepackage{float}
\usepackage{xcolor}
\usepackage[super]{nth}
\usepackage{booktabs}
\usepackage{bbm}
\usepackage{dsfont}
\definecolor{blue-violet}{rgb}{0.54, 0.17, 0.89}
\usepackage[pagebackref,breaklinks,colorlinks]{hyperref}

\usepackage[capitalize]{cleveref}
\crefname{section}{Sec.}{Secs.}
\Crefname{section}{Section}{Sections}
\Crefname{table}{Table}{Tables}
\crefname{table}{Tab.}{Tabs.}

\def\cvprPaperID{2683} % *** Enter the CVPR Paper ID here
\def\confName{CVPR}
\def\confYear{2023}

\begin{document}

%%%%%%%%% TITLE - PLEASE UPDATE
\title{Dream3D: Zero-Shot Text-to-3D Synthesis Using 3D Shape Prior and Text-to-Image Diffusion Models}

\author{
Jiale Xu\textsuperscript{\rm 1,3}\footnotemark[1] \quad 
Xintao Wang\textsuperscript{\rm 1}\footnotemark[2] \quad
Weihao Cheng\textsuperscript{\rm 1} \quad
Yan-Pei Cao\textsuperscript{\rm 1} \\
Ying Shan\textsuperscript{\rm 1} \quad
Xiaohu Qie\textsuperscript{\rm 2} \quad
Shenghua Gao\textsuperscript{\rm 3,4,5}\footnotemark[2]\\
\textsuperscript{\rm 1}ARC Lab, \textsuperscript{\rm 2}Tencent PCG \quad
\textsuperscript{\rm 3}ShanghaiTech University \\
\textsuperscript{\rm 4}Shanghai Engineering Research Center of Intelligent Vision and Imaging \\
\textsuperscript{\rm 5}Shanghai Engineering Research Center of Energy Efficient and Custom AI IC \\
}
\twocolumn[{%
\maketitle
\vspace{-1.2cm}
\begin{figure}[H]
\hsize=\textwidth % cvpr 需要
\centering
\includegraphics[width=\textwidth]{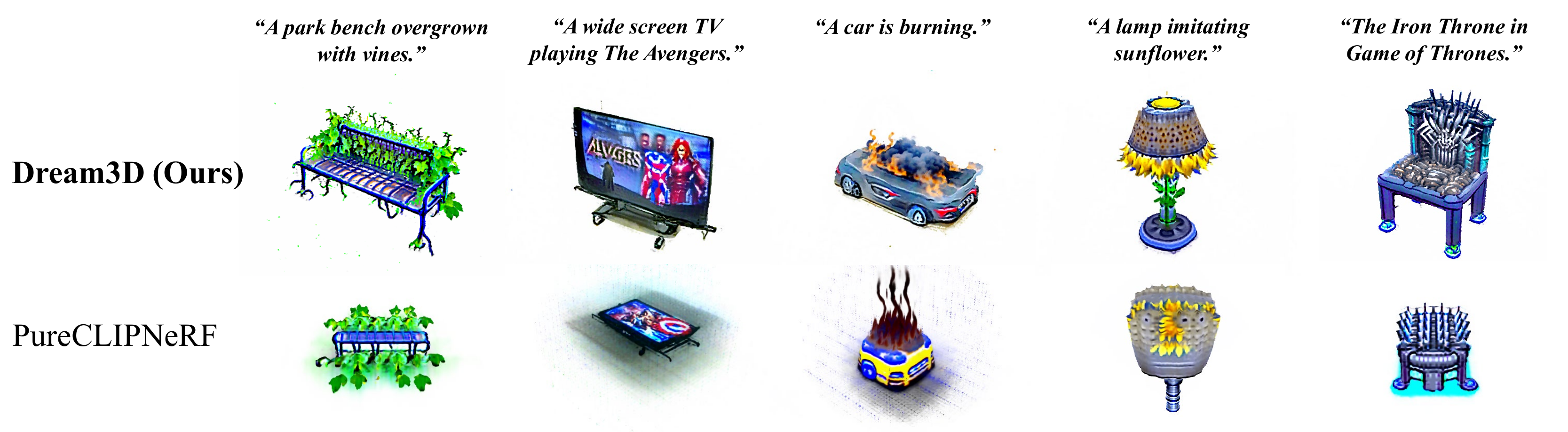}
   \caption{By utilizing 3D shape priors and powerful text-to-image diffusion models, our Dream3D can generate 3D content that exhibits superior visual quality and shape accuracy in accordance with the text prompt when compared to PureCLIPNeRF~\cite{lee2022understanding}.}
   \label{fig:teaser}
\end{figure}
}]

\renewcommand{\thefootnote}{\fnsymbol{footnote}}
\footnotetext[1]{Work done during an internship at ARC Lab, Tencent PCG.}
\footnotetext[2]{Corresponding Author.}

%%%%%%%%% ABSTRACT
\begin{abstract}
\vspace{-0.5cm}
% Recent CLIP-guided 3D optimization methods \eg, DreamFields~\cite{Jain_2022_CVPR} and PureCLIPNeRF~\cite{lee2022understanding} achieve great success in zero-shot text-guided 3D synthesis. However, due to the scratch training and random initialization without any prior knowledge, these methods usually fail to generate accurate and faithful 3D structures that conform to the corresponding text.
% In this paper, we make the first attempt to introduce the explicit 3D shape prior to CLIP-guided 3D optimization methods. Specifically, we first generate a high-quality 3D shape from the input text in the text-to-shape stage as a 3D shape prior. We then utilize it as the initialization of a neural radiance field and optimize with the full prompt.
% For the challenging text-to-shape generation process, we present a simple yet effective approach that directly bridges the text and image modalities with a powerful text-to-image diffusion model. To narrow the style domain gap between the images synthesized by the text-to-image diffusion model and shape renderings used to train the image-to-shape generator, we further propose to jointly optimize a learnable text prompt and fine-tune the text-to-image diffusion model for rendering-style image generation.
% Our method, namely, Dream3D, is capable of generating imaginative 3D contents with better visual quality and shape accuracy than state-of-the-art methods. Our Project page is at \url{https://bluestyle97.github.io/dream3d/}.

Recent CLIP-guided 3D optimization methods, such as DreamFields~\cite{Jain_2022_CVPR} and PureCLIPNeRF~\cite{lee2022understanding}, have achieved impressive results in zero-shot text-to-3D synthesis. However, due to scratch training and random initialization without prior knowledge, these methods often fail to generate accurate and faithful 3D structures that conform to the input text. In this paper, we make the first attempt to introduce explicit 3D shape priors into the CLIP-guided 3D optimization process. Specifically, we first generate a high-quality 3D shape from the input text in the text-to-shape stage as a 3D shape prior. We then use it as the initialization of a neural radiance field and optimize it with the full prompt. To address the challenging text-to-shape generation task, we present a simple yet effective approach that directly bridges the text and image modalities with a powerful text-to-image diffusion model. To narrow the style domain gap between the images synthesized by the text-to-image diffusion model and shape renderings used to train the image-to-shape generator, we further propose to jointly optimize a learnable text prompt and fine-tune the text-to-image diffusion model for rendering-style image generation. Our method, Dream3D, is capable of generating imaginative 3D content with superior visual quality and shape accuracy compared to state-of-the-art methods. Our project page is at \url{https://bluestyle97.github.io/dream3d/}.

\end{abstract}

%%%%%%%%% BODY TEXT
\section{Introduction}
\label{sec:intro}

% Text-guided 3D synthesis aims to generate 3D content consistent with an input text, which has the potential to benefit a broad range of applications, \eg, animations, games, and virtual reality. Modern zero-shot text-to-image models~\cite{ramesh2021zero, nichol2021glide, ramesh2022hierarchical, saharia2022photorealistic, Rombach_2022_CVPR} have achieved unprecedented success and can synthesize diverse, high-fidelity, and imaginative images from various text prompts. However, it is challenging to replicate the growth in text-to-image synthesis to the text-to-3D synthesis task since collecting a large-scale paired text-3D dataset is unfeasible.

Text-to-3D synthesis endeavors to create 3D content that is coherent with an input text, which has the potential to benefit a wide range of applications such as animations, games, and virtual reality. Recently developed zero-shot text-to-image models~\cite{ramesh2021zero, nichol2021glide, ramesh2022hierarchical, saharia2022photorealistic, Rombach_2022_CVPR} have made remarkable progress and can generate diverse, high-fidelity, and imaginative images from various text prompts. However, extending this success to the text-to-3D synthesis task is challenging because it is not practically feasible to collect a comprehensive paired text-3D dataset.

% Zero-shot text-to-3D synthesis~\cite{Sanghi_2022_CVPR, Jain_2022_CVPR, khalid2022clipmesh, poole2022dreamfusion, lee2022understanding}, on the contrary, usually leverages powerful vision-language models, \eg, CLIP~\cite{wang2022clip}, to eliminate the demand for paired data, and thus is more attractive. Typically, there are two categories. \textbf{1)} CLIP-based generative models~\cite{Sanghi_2022_CVPR, liu2022iss}. CLIP-Forge~\cite{Sanghi_2022_CVPR} is a representative method, which uses images as an intermediate bridge. It trains a mapper from the CLIP image embeddings of ShapeNet renderings to the shape embeddings of a 3D shape generator, then switches to CLIP text embedding as the mapper's input at test time. \textbf{2)} CLIP-guided 3D optimization methods, \eg, DreamFields~\cite{Jain_2022_CVPR} and PureCLIPNeRF~\cite{lee2022understanding}. They continuously optimize the CLIP similarity loss between the text prompt and rendered images of a 3D scene representation, \eg, neural radiance fields~\cite{mildenhall2020nerf}. The first category heavily relies on 3D shape generators trained on limited 3D shapes and seldom has the capacity to adjust its shape structures, while the second category has more creative freedom with the ``dreaming ability'' to generate diverse shape structures and textures. 

Zero-shot text-to-3D synthesis~\cite{Sanghi_2022_CVPR, Jain_2022_CVPR, khalid2022clipmesh, poole2022dreamfusion, lee2022understanding}, which eliminates the need for paired data, is an attractive approach that typically relies on powerful vision-language models such as CLIP~\cite{wang2022clip}. There are two main categories of this approach. \textbf{1)} CLIP-based generative models, such as CLIP-Forge~\cite{Sanghi_2022_CVPR}. They utilize images as an intermediate bridge and train a mapper from the CLIP image embeddings of ShapeNet renderings to the shape embeddings of a 3D shape generator, then switch to the CLIP text embedding as the input at test time. \textbf{2)} CLIP-guided 3D optimization methods, such as DreamFields~\cite{Jain_2022_CVPR} and PureCLIPNeRF~\cite{lee2022understanding}. They continuously optimize the CLIP similarity loss between a text prompt and rendered images of a 3D scene representation, such as neural radiance fields~\cite{mildenhall2020nerf,Barron2022MipNeRF3U,pumarola2021d,liu2022devrf}. While the first category heavily relies on 3D shape generators trained on limited 3D shapes and seldom has the capacity to adjust its shape structures, the second category has more creative freedom with the ``dreaming ability'' to generate diverse shape structures and textures. 

% We start to develop our method from CLIP-guided 3D optimization methods. Though these methods can produce impressive results, we observe that they usually fail to generate accurate and faithful 3D structures that conform to the corresponding texts (Fig.~\ref{fig:teaser}, $2^{nd}$ row). Due to the scratch training and random initialization without any prior knowledge, these methods tend to generate highly-unconstrained ``adversarial contents'' that have high CLIP scores but low visual quality. To overcome the aforementioned issue and synthesize more faithful 3D contents, we propose to first generate a high-quality 3D shape from the input text, and then leverage it as an explicit \textbf{“3D shape prior”} to the CLIP-guided 3D optimization process. Specifically, in the \textit{text-to-shape}\footnote{We use the term ``shape'' to denote the 3D geometric models \textit{without textures} throughout the paper, while some works~\cite{chen2018text2shape,Liu_2022_CVPR} also use this term for textured 3D models.} stage, we first synthesize a 3D shape without textures of the main common object in the text prompt. We then utilize it as the \textit{initialization} of a voxel-based neural radiance field and then optimize with the full prompt.

We develop our method building upon CLIP-guided 3D optimization methods. Although these methods can produce remarkable outcomes, they typically fail to create precise and accurate 3D structures that conform to the input text (\cref{fig:teaser}, \nth{2} row)). Due to the scratch training and random initialization without any prior knowledge, these methods tend to generate highly-unconstrained ``adversarial contents" that have high CLIP scores but low visual quality. To address this issue and synthesize more faithful 3D contents, we suggest generating a high-quality 3D shape from the input text first and then using it as an explicit \textbf{``3D shape prior"} in the CLIP-guided 3D optimization process. In the \textit{text-to-shape}\footnote{Throughout this paper, we use the term ``shape" to refer to 3D geometric models \textit{without textures}, while some works~\cite{chen2018text2shape,Liu_2022_CVPR} also use this term for textured 3D models.} stage, we begin by synthesizing a 3D shape without textures of the main common object in the text prompt. We then use it as the \textit{initialization} of a voxel-based neural radiance field and optimize it with the full prompt.

% The \textit{text-to-shape} generation itself is a challenging task. Previous methods~\cite{Sanghi_2022_CVPR, Jain_2022_CVPR} are often trained on images and tested with texts, and use CLIP to bridge the two modalities. Such a design leads to a mismatching problem due to the gap between the CLIP text and image embedding spaces. Moreover, existing methods are incapable of generating high-quality 3D shapes. In this work, we  propose to directly bridge the text and image modalities with a powerful text-to-image diffusion model, \ie, Stable Diffusion~\cite{Rombach_2022_CVPR}. Specifically, we adopt the \textit{text-to-image} diffusion model to synthesize an image from the input text, and then feed the image into an \textit{image-to-shape} generator to produce high-quality 3D shapes. Due to the same procedure in both training and testing, the mismatching problem can be largely reduced. However, it still suffers from the style domain gap between the images synthesized by the text-to-image diffusion model and the shape renderings used to train the image-to-shape generator. Inspired by recent work on controllable text-to-image synthesis~\cite{gal2022image,ruiz2022dreambooth}, we propose to jointly optimize a learnable text prompt and fine-tune the Stable Diffusion to address this domain gap. The fine-tuned Stable Diffusion can synthesize images in the style of shape renderings that we used to train the image-to-shape module reliably, without suffering from the domain gap.

The \textit{text-to-shape} generation itself is a challenging task. Previous methods~\cite{Sanghi_2022_CVPR, Jain_2022_CVPR} are often trained on images and tested with texts, and use CLIP to bridge the two modalities. However, this approach leads to a mismatching problem due to the gap between the CLIP text and image embedding spaces. Additionally, existing methods cannot produce high-quality 3D shapes. In this work, we  propose to directly bridge the text and image modalities with a powerful text-to-image diffusion model, \ie, Stable Diffusion~\cite{Rombach_2022_CVPR}. We use the \textit{text-to-image} diffusion model to synthesize an image from the input text and then feed the image into an \textit{image-to-shape} generator to produce high-quality 3D shapes. Since we use the same procedure in both training and testing, the mismatching problem is largely reduced. However, there is still a style domain gap between the images synthesized by Stable Diffusion and the shape renderings used to train the image-to-shape generator. Inspired by recent work on controllable text-to-image synthesis~\cite{gal2022image,ruiz2022dreambooth}, we propose to jointly optimize a learnable text prompt and fine-tune the Stable Diffusion to address this domain gap. The fine-tuned Stable Diffusion can reliably synthesize images in the style of shape renderings used to train the image-to-shape module without suffering from the domain gap.

% To summarize,
% \textbf{1)} We make the first attempt to introduce the explicit \textbf{3D shape prior} into CLIP-guided 3D optimization methods. The proposed method can generate more accurate and high-quality \textbf{3D shapes} conforming  to the corresponding text, while still enjoying the \textbf{``dreaming'' ability} of generating diverse shape structures and textures (Fig.~\ref{fig:teaser}, $1^{st}$ row). Therefore, we name our method ``Dream3D'' as it has both strengths.
% \textbf{2)} For text-to-shape generation, we present a simple yet effective approach that directly bridges the text and image modalities with a powerful text-to-image diffusion model. In order to narrow the style domain gap between the synthesized images and shape renderings, we further propose to jointly optimize a learnable text prompt and fine-tune a powerful text-to-image diffusion model for rendering-style image generation.
% \textbf{3)} Our Dream3D is capable of generating imaginative 3D contents with better visual quality and shape accuracy than state-of-the-art methods. Moreover, our text-to-shape pipeline can also generate 3D shapes with higher quality than previous work.

To summarize, 
\textbf{1)} We make the first attempt to introduce the explicit \textbf{3D shape prior} into CLIP-guided 3D optimization methods. The proposed method can generate more accurate and high-quality \textbf{3D shapes} conforming  to the corresponding text, while still enjoying the \textbf{``dreaming'' ability} of generating diverse shape structures and textures (\cref{fig:teaser}, \nth{1} row). Therefore, we name our method ``Dream3D'' as it has both strengths.
\textbf{2)} Regarding text-to-shape generation, we present a straightforward yet effective approach that directly connects the text and image modalities using a powerful text-to-image diffusion model. To narrow the style domain gap between the synthesized images and shape renderings, we further propose to jointly optimize a learnable text prompt and fine-tune the text-to-image diffusion model for rendering-style image generation.
\textbf{3)} Our Dream3D can generate imaginative 3D content with better visual quality and shape accuracy than state-of-the-art methods. Additionally, our text-to-shape pipeline can produce 3D shapes of higher quality than previous work.

\section{Related Work}
\label{sec:related}

\noindent\textbf{3D Shape Generation.} 
Generative models for 3D shapes have been extensively studied in recent years. It is more challenging than 2D image generation due to the expensive 3D data collection and the complexity of 3D shapes. Various 3D generators employ different shape representations, \eg, voxel grids~\cite{Tatarchenko_2017_ICCV, li2017grass,cao2018learning}, point clouds~\cite{Yang_2019_ICCV, cai2020learning, Zhou_2021_ICCV, zeng2022lion,xiang2022snowflake}, meshes~\cite{gao2019sdm, nash2020polygen, NEURIPS2020_1349b36b, gao2020tmnet}, and implicit fields~\cite{Chen_2019_CVPR, Wu_2020_CVPR, Luo_2021_ICCV, zheng2022sdfstylegan, wu2022learning}. These generators are trained to model the distribution of shape geometry (and optionally, texture) from a collection of 3D shapes. Some methods~\cite{pavllo2020convolutional, Pavllo_2021_ICCV, gao2022get3d, Chan_2021_CVPR, schwarz2020graf, Chan_2022_CVPR, Or-El_2022_CVPR} attempt to learn a 3D generator using only 2D image supervision. These methods incorporate explicit 3D representations, such as meshes~\cite{pavllo2020convolutional, Pavllo_2021_ICCV, gao2022get3d} and neural radiance fields~\cite{Chan_2021_CVPR, schwarz2020graf, Chan_2022_CVPR, Or-El_2022_CVPR}, along with surface or volume-based differentiable rendering techniques~\cite{mildenhall2020nerf, Kato_2018_CVPR, Jiang_2020_CVPR, Niemeyer_2020_CVPR}, to enable the learning of 3D awareness from images.

\begin{figure*}[t]
  \centering
   \includegraphics[width=\linewidth]{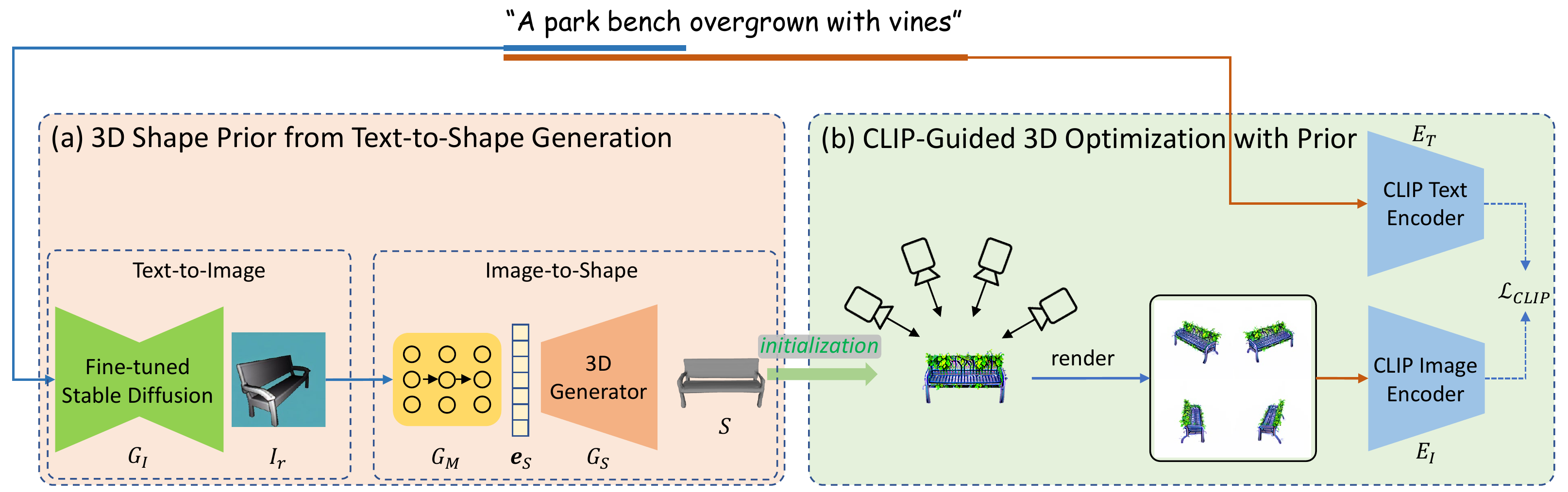}
   \vspace{-6mm}
   \caption{Overview of our text-to-3D synthesis framework. (a) In the first text-to-shape stage, a fine-tuned Stable Diffusion $G_I$ is employed to synthesize a rendering-style image $\boldsymbol{I}_r$ from the input text prompt $y$. This image is then used to generate a latent shape embedding $\boldsymbol{e}_S$ with the assistance of a shape embedding generation network $G_M$. Finally, the high-quality 3D shape generator $G_S$ leverages $\boldsymbol{e}_S$ to produce a 3D shape $S$, which is used as an explicit 3D shape prior. (b) In the second optimization stage, the 3D shape prior $S$ is utilized to initialize a neural radiance field, which is further optimized with CLIP guidance to synthesize 3D content that is consistent with the input text prompt $y$.}
   \label{fig:framework}
   \vspace{-4mm}
\end{figure*}

\noindent\textbf{Text-to-Image.} 
Previous studies in text-to-image synthesis~\cite{reed2016generative, Zhang_2017_ICCV, Xu_2018_CVPR, Qiao_2019_CVPR} have focused mainly on domain-specific datasets and utilized GANs~\cite{goodfellow2020generative}. However, recent advances in scalable generative architectures and large-scale text-image datasets~\cite{schuhmann2022laion} have enabled unprecedented performance in zero-shot text-to-image synthesis. DALL$\cdot$E~\cite{ramesh2021zero} and GLIDE~\cite{nichol2021glide}, as pioneering works, employ auto-regressive model~\cite{Esser_2021_CVPR} and diffusion model~\cite{song2020denoising, ho2020denoising} as their architectures, respectively. DALL$\cdot$E 2~\cite{ramesh2022hierarchical} utilizes a diffusion prior network to translate CLIP text embeddings to CLIP image embeddings, and an unCLIP module to synthesize images from CLIP image embeddings. In Imagen~\cite{saharia2022photorealistic} and Stable Diffusion~\cite{Rombach_2022_CVPR}, a large pre-trained text encoder is employed to guide the sampling process of a diffusion model in pixel space and latent space, respectively.

\noindent\textbf{Zero-Shot Text-to-3D.} 
Zero-shot text-to-3D generation techniques~\cite{Sanghi_2022_CVPR, Jain_2022_CVPR, Michel_2022_CVPR, khalid2022clipmesh, poole2022dreamfusion, lee2022understanding} exploit the joint text-image modeling capability of pre-trained vision-language models such as CLIP~\cite{radford2021learning} to obviate the need for paired text-3D data. CLIP-Forge~\cite{Sanghi_2022_CVPR} trains a normalizing flow~\cite{rezende2015variational, dinh2017density} model to convert CLIP image embeddings to VAE~\cite{kingma2013auto} shape embeddings, and switches the input to CLIP text embeddings at the inference time. ISS~\cite{liu2022iss} trains a mapper to map the CLIP image embedding into the latent shape code of a pre-trained single-view reconstruction (SVR) network~\cite{Niemeyer_2020_CVPR}, which is then fine-tuned by taking the CLIP text embedding as input. DreamFields~\cite{Jain_2022_CVPR} and CLIP-Mesh~\cite{khalid2022clipmesh} are pioneering works that explore zero-shot 3D content creation using only CLIP guidance. The former optimizes a randomly-initialized NeRF, while the latter optimizes a spherical template mesh as well as random texture and normal maps. PureCLIPNeRF~\cite{lee2022understanding} enhances DreamFields with grid-based representation~\cite{Sun_2022_CVPR_DVGO} and more diverse image augmentations. Recently, DreamFusion~\cite{poole2022dreamfusion} has gained popularity in the research community due to its impressive results. Powered by a strong text-to-image model, Imagen~\cite{saharia2022photorealistic}, it can generate high-fidelity 3D objects using the score distillation loss.

%------------------------------------------------------------------------
\section{Method}
\label{sec:method}

% Given an input text prompt $y$, we aim to generate 3D content consistent with the description of $y$. As shown in \cref{fig:framework}, our text-guided 3D synthesis framework consists of two stages. In the first stage (Sec.~\ref{sec:text2shape}), we acquire an explicit 3D shape prior $S$ with a text-guided 3D shape generation process. The text-guided shape generation process is further split into a text-to-image phase using a fine-tuned Stable Diffusion model~\cite{Rombach_2022_CVPR} $G_I$ and an image-to-shape phase using a shape embedding generation network $G_M$ together with a high-quality 3D shape generator $G_S$. In the second stage, we leverage the 3D prior $S$ to initialize a neural radiance field~\cite{mildenhall2020nerf} and optimize it with the CLIP~\cite{radford2021learning} guidance to generate the 3D content. Our framework only requires a collection of textureless 3D shapes without any text labels for training the 3D generator $G_S$, and the fine-tuning process of $G_I$ converges very fast.

Our objective is to generate 3D content that aligns with the given input text prompt $y$. As illustrated in \cref{fig:framework}, our framework for text-guided 3D synthesis comprises two stages. In the first stage (\cref{sec:text2shape}), we obtain an explicit 3D shape prior $S$ using a text-guided 3D shape generation process. The text-guided shape generation process involves a text-to-image phase that employs a fine-tuned Stable Diffusion model~\cite{Rombach_2022_CVPR} $G_I$ (\cref{sec:stylize}), and an image-to-shape phase that employs a shape embedding generation network $G_M$ and a high-quality 3D shape generator $G_S$. In the second stage  (\cref{sec:clip_guided_optimization}), we utilize the 3D prior $S$ to initialize a neural radiance field~\cite{mildenhall2020nerf}, and optimize it with CLIP~\cite{radford2021learning} guidance to generate the 3D content. Our framework only requires a collection of textureless 3D shapes without any text labels to train the 3D generator $G_S$, and the fine-tuning process of $G_I$ converges rapidly.

\subsection{CLIP-Guided 3D Optimization with explicit 3D Shape Prior}
\label{sec:clip_guided_optimization}

\noindent\textbf{Background: 3D Optimization with CLIP Guidance.} 
CLIP~\cite{radford2021learning} is a powerful vision-language model that comprises a text encoder $E_T$, and an image encoder $E_I$. By maximizing the cosine similarity between the text embedding and the image embedding encoded by $E_T$ and $E_I$ respectively on a large-scale paired text-image dataset, CLIP aligns the text and image modalities in a shared latent embedding space.

Prior research~\cite{Jain_2022_CVPR, khalid2022clipmesh, lee2022understanding} leverages the capability of CLIP~\cite{radford2021learning} to generate 3D contents from text. Starting from a randomly-initialized 3D representation parameterized by $\theta$, they render images from multiple viewpoints and optimize $\theta$ by minimizing the CLIP similarity loss between the rendered image $\mathcal{R}(\boldsymbol{v}_i;\theta)$ and the text prompt $y$: 
\begin{equation}
    \mathcal{L}_{\text{CLIP}} = -E_I(\mathcal{R}(\theta;\boldsymbol{v}_i))^TE_T(y),
    \label{eq:clip_loss}
\end{equation}
where $\mathcal{R}$ denotes the rendering process, and $\boldsymbol{v}_i$ denotes the rendering viewpoint at the $i$-th optimization step.
Specifically, DreamFields~\cite{Jain_2022_CVPR} and PureCLIPNeRF~\cite{lee2022understanding} employ neural radiance fields~\cite{mildenhall2020nerf} (NeRF) as the 3D representation $\theta$, while CLIP-Mesh~\cite{khalid2022clipmesh} uses a spherical template mesh with associated texture and normal maps.

\noindent\textbf{Observations and Motivations.} 
Though these CLIP-guided optimization methods can generate impressive results, we observe that they often fall short in producing precise and detailed 3D structures that accurately match the text description. As depicted in \cref{fig:teaser}, we employ these methods to create 3D content featuring common objects, but the outcomes exhibited distortion artifacts and appeared unusual, adversely affecting their visual quality and hindering their use in real-world applications.

We attribute the failure of previous works to generate accurate and realistic objects to two main factors:
\textbf{(i)} The optimization process begins with a \textit{randomly-initialized 3D representation} lacking any explicit 3D shape prior, making it very challenging for the models to conjure up the scene from scratch.
\textbf{(ii)} The CLIP loss in \cref{eq:clip_loss} prioritizes global consistency between the rendered image and the text prompt, rather than offering robust and precise guidance on the synthesized 3D structure. As a result, the optimization output is significantly unconstrained.

\noindent\textbf{Optimization with 3D Shape Prior as Initialization.}
To address the aforementioned issue and generate more faithful 3D content, we propose to use a text-to-shape generation process to create a \textit{high-quality 3D shape} $S$ from the input text prompt $y$. Subsequently, we use it as an explicit ``3D shape prior" to initialize the CLIP-guided 3D optimization process. As illustrated in \cref{fig:framework}, for the text prompt \textit{``a park bench overgrown with vines"}, we first synthesize \textit{``a park bench"} without textures in the text-to-shape stage. We then use it as the initialization of a neural radiance field and optimize it with the full prompt, following previous works.

Our optimization process utilizes DVGO~\cite{Sun_2022_CVPR_DVGO}, an efficient NeRF variant that represents NeRF using a density voxel grid $\boldsymbol{V}_{\text{density}} \in \mathbb{R}^{N_x\times N_y\times N_z}$ and a shallow color MLP $f_{\text{rgb}}$. We start by taking a 3D shape $S$ generated by the text-to-shape process, represented as an SDF grid $\tilde{\boldsymbol{V}}_{\text{sdf}} \in \mathbb{R}^{N_x\times N_y\times N_z}$, and transform it into the density voxel grid $\boldsymbol{V}_{\text{density}}$ using the following equations~\cite{yariv2021volume, or2022stylesdf, Sun_2022_CVPR_DVGO}:
\begin{subequations}
\begin{align}
    \boldsymbol{\Sigma} &= \frac{1}{\beta}\operatorname{sigmoid}\left(-\frac{\tilde{\boldsymbol{V}}_{\text{sdf}}}{\beta}\right),
    \label{eq:sdf2density1} \\
    \boldsymbol{V}_{\text{density}} &= \max(0, \operatorname{softplus}^{-1}(\boldsymbol{\Sigma}))
    \label{eq:sdf2density2}
\end{align}
\end{subequations}
Here, $\operatorname{sigmoid}(x)=1/(1+e^{-x})$ and $\operatorname{softplus}^{-1}(x)=\log(e^x-1)$. \cref{eq:sdf2density1} converts SDF values to density for volume rendering, where $\beta>0$ is a hyper-parameter controlling the sharpness of the shape boundary (smaller $\beta$ leads to sharper shape boundary, $\beta=0.05$ in our experiments). \cref{eq:sdf2density2} transforms the density into pre-activated density. To ensure that the distribution of the accumulated transmittance is the same as DVGO, we clamp the minimum value of the density outside the shape prior as $0$.

With the density grid $\boldsymbol{V}_{\text{density}}$ initialized by the 3D shape $S$ and the color MLP $f_{\text{rgb}}$ initialized randomly, we render image $\mathcal{R} (\boldsymbol{V}_{\text{density}},f_{\text{rgb}};\boldsymbol{v}_i)$ from viewpoint $\boldsymbol{v}_i$ and optimize $\theta=(\boldsymbol{V}_{\text{density}},f_{\text{rgb}})$ with the CLIP loss in \cref{eq:clip_loss}. Following DreamFields~\cite{Jain_2022_CVPR} and PureCLIPNeRF~\cite{lee2022understanding}, we perform background augmentations for the rendered images and leverage the transmittance loss introduced by~\cite{Jain_2022_CVPR} to reduce noise and spurious density. Besides, since CLIP loss cannot provide accurate geometrical supervision, the 3D shape prior may be gradually disturbed and ``forgotten", thus we also adopt a shape-prior-preserving loss to preserve the global structure of the 3D shape prior:
\begin{equation}
    \mathcal{L}_{\text{prior}} = -\sum_{(x,y,z)} \mathds{1}(\tilde{\boldsymbol{V}}_{\text{sdf}}<0)\cdot\operatorname{alpha}(\boldsymbol{V}_{\text{density}}),
    \label{eq:prior_loss}
\end{equation}
where $\mathds{1}(\cdot)$ is the indicator function, and $\operatorname{alpha}(\cdot)$ transforms the density into the opacity representing the probability of termination at each position in volume rendering.

By initializing NeRF with an explicit 3D shape prior, we give extra knowledge on how the 3D content should look like and prevent the model from imagining from scratch and generating ``adversarial contents" that have high CLIP scores but low visual quality. Based on the initialization, the CLIP-guided optimization further provides flexibility and is able to synthesize \textit{more diverse structures and textures}.

% \subsection{CLIP-Guided Optimization with Priors}
% Beyond generating 3D shapes from text prompts, we hope to synthesize more diverse and imaginative 3D content to facilitate real applications such as virtual reality. Previous work on 3D content synthesis, \eg, DreamFields~\cite{Jain_2022_CVPR}, CLIP-Mesh~\cite{khalid2022clipmesh}, and PureCLIPNeRF~\cite{lee2022understanding} directly optimizes a scene representation from scratch by maximizing the CLIP similarity between rendered images and text prompt. Due to the lack of 3D priors, these approaches tend to synthesize irregular and unrealistic 3D structures.

% We represent a 3D scene with an efficient NeRF~\cite{mildenhall2020nerf} representation, \ie, DVGO~\cite{Sun_2022_CVPR_DVGO}, which consists of a density voxel grid for geometry and a shallow MLP network for color prediction. Since the 3D shape synthesized by the image-to-shape generator is represented with an SDF grid $\boldsymbol{V}_{sdf}\in \mathbb{R}^{N\times N\times N}$, we first transform it into a density voxel grid $\boldsymbol{V}_{density}$ using the following equation:
% \begin{equation}
%     \boldsymbol{V}_{density} = \operatorname{softplus}^{-1}(\frac{1}{\beta}\operatorname{sigmoid}(-\frac{\boldsymbol{V}_{sdf}}{\beta})
%     \label{eq:sdf2density}
% \end{equation}
% where sig $\beta>0$ is a hyper-parameter controlling the sharpness of the shape boundary. $\beta$ values approaching 0 represent a solid and sharp shape boundary, whereas larger $\beta$ values indicate a more fluffy shape boundary.

%\subsection{Text-to-Shape Generation as 3D Prior}
\subsection{Stable-Diffusion-Assisted Text-to-Shape Generation as 3D Shape Prior}
\label{sec:text2shape}

To obtain the 3D shape prior, a text-guided shape generation scheme is required, which is a challenging task due to the lack of paired text-shape datasets. Previous approaches~\cite{Sanghi_2022_CVPR, liu2022iss} typically first train an image-to-shape model using \textit{rendered} images, and then bridge the text and image modalities using the CLIP embedding space. 

CLIP-Forge~\cite{Sanghi_2022_CVPR} trains a normalizing flow network to map CLIP image embeddings of shape renderings to latent embeddings of a volumetric shape auto-encoder, and at test time, it switches to CLIP text embeddings as input. However, the shape auto-encoder has difficulty in generating high-quality and diverse 3D shapes, and directly feeding CLIP text embeddings to the flow network trained on CLIP image embeddings suffers from the gap between the CLIP text and image embedding spaces. ISS~\cite{liu2022iss} trains a mapper network to map CLIP image embeddings of shape renderings to the latent space of a pre-trained single-view reconstruction (SVR) model, and fine-tunes the mapper at test time by maximizing the CLIP similarity between the text prompt and the images rendered from synthesized shapes. While the test-time fine-tuning alleviates the gap between the CLIP text and image embeddings, it is cumbersome to fine-tune the mapper for each text prompt.

In contrast to the aforementioned methods that connect the text and image modalities in the CLIP embedding space, we use a powerful text-to-image diffusion model to directly bridge the two modalities. Specifically, we first synthesize an image from the input text and then feed it into an image-to-shape module to generate a \textit{high-quality} 3D shape. This pipeline is more concise and naturally eliminates the gap between CLIP text and image embeddings. However, it introduces a new domain gap between the images generated by the text-to-image diffusion model and the shape renderings used to train the image-to-shape module. We will introduce a novel technique to alleviate this gap in \cref{sec:stylize}.

\noindent\textbf{Text-to-Image Diffusion Model.} 
Diffusion models~\cite{ho2020denoising, song2020denoising} are generative models trained to reverse a diffusion process. The diffusion process begins with a sample from the data distribution, $\boldsymbol{x}_0 \sim q\left(\boldsymbol{x}_0\right)$, which is gradually corrupted by Gaussian noise over $T$ timesteps: $\boldsymbol{x}_t=\sqrt{\alpha_t} \boldsymbol{x}_{t-1}+\sqrt{1-\alpha_t} \boldsymbol{\epsilon}_{t-1},t=1,2,\ldots, T$, where $\alpha_t$ defines the noise level and $\boldsymbol{\epsilon}_{t-1}$ denotes the noise added at timestep $t-1$.  To reverse this process, a denoising network $\epsilon_{\theta}$ is trained to estimate the added noise at each timestep. During inference, samples can be generated by iteratively denoising pure Gaussian noise. Text-to-image diffusion models further condition the denoising process on texts. Given a text prompt $y$ and a text encoder $c_\theta$, the training objective is:
\begin{equation}
\mathcal{L}_{\text{diffusion}}=\mathbb{E}_{\boldsymbol{x}_0, t, \boldsymbol{\epsilon}_t, y}\left\|\boldsymbol{\epsilon}_t-\epsilon_\theta\left(\boldsymbol{x}_t, t, c_\theta(y)\right)\right\|_2^2
\label{eq:diffusion_loss_conditional}
\end{equation}

In this work, we use Stable Diffusion\footnote{\url{https://github.com/CompVis/stable-diffusion}}~\cite{Rombach_2022_CVPR}, an open-source text-to-image diffusion model, which employs a CLIP ViT-L/14 text encoder as $c_\theta$ and is trained on the large-scale LAION-5B dataset~\cite{schuhmann2022laion}. Stable Diffusion is known for its ability to generate diverse and imaginative images in various styles from heterogeneous text prompts.

% \noindent\textbf{High-quality 3D generator.} 
% High-quality 3D shapes are highly desired for providing a more precise initialization for optimization. We adopt a state-of-the-art 3D generative model architecture, \ie, SDF-StyleGAN~\cite{zheng2022sdfstylegan}, to provide high-quality 3D priors. SDF-StyleGAN is a StyleGAN2-like generative architecture that maps a random noise $\boldsymbol{z} \sim \mathcal{N}(\boldsymbol{0}, \boldsymbol{I})$ to a latent shape embedding $\boldsymbol{e}_S\in \mathcal{W}$ and synthesizes a 3D feature volume $\boldsymbol{F}_V$ - an implicit shape representation. We can query the SDF value at arbitrary position $\boldsymbol{x}$ by feeding the interpolated feature from $\boldsymbol{F}_V$ at $\boldsymbol{x}$ into a jointly trained MLP network. Different from the original SDF-StyleGAN that trains one network for one shape category, we improve it to train \textit{one} 3D shape generator $G_S$ on 13 categories of the ShapeNet~\cite{chang2015shapenet} dataset, so that the text-to-shape process has more flexibility.

\noindent\textbf{High-quality 3D generator.} 
To provide a more precise initialization for optimization, high-quality 3D shapes are highly desirable. In this work, we utilize the SDF-StyleGAN~\cite{zheng2022sdfstylegan}, a state-of-the-art 3D generative model, to generate high-quality 3D priors. SDF-StyleGAN is a StyleGAN2-like architecture that maps a random noise $\boldsymbol{z} \sim \mathcal{N}(\boldsymbol{0}, \boldsymbol{I})$ to a latent shape embedding $\boldsymbol{e}_S\in \mathcal{W}$ and synthesizes a 3D feature volume $\boldsymbol{F}_V$, which is an implicit shape representation. We can query the SDF value at arbitrary position $\boldsymbol{x}$ by feeding the interpolated feature from $\boldsymbol{F}_V$ at $\boldsymbol{x}$ into a jointly trained MLP. We improve upon the original SDF-StyleGAN, which trains one network for each shape category, by training a \textit{single} 3D shape generator $G_S$ on the 13 categories of ShapeNet~\cite{chang2015shapenet}. This modification provides greater flexibility in the text-to-shape process.

\noindent\textbf{Shape Embedding Mapping Network.} 
To bridge the image and shape modalities, we further train a shape embedding mapping network $G_M$. Firstly, we utilize $G_S$ to generate a large set of 3D shapes $\{S^i\}_{i=1}^N$ and shape embeddings $\{\boldsymbol{e}_S^i\}_{i=1}^N$. Then, we render $S^i$ from $K$ viewpoints to obtain shape renderings $\{\boldsymbol{I}_r^{j}\}_{j=1}^{NK}$ and the corresponding image embeddings $\{\boldsymbol{e}_I^{j}\}_{j=1}^{NK}$ with the CLIP image encoder $E_I$, forming a paired image-shape embedding dataset $\{(\boldsymbol{e}_I^{j}, \boldsymbol{e}_S^j)\}_{j=1}^{NK}$. Finally, we use this dataset to train a conditional diffusion model $G_M$ which can synthesize shape embeddings from image embeddings of shape renderings.

To prepare the dataset for training $G_M$, we generate $N=64000$ shapes using $G_S$ and render $K=24$ views for each shape. The ranges of azimuth and elevation angles of the rendered views are $\left[-90^{\circ}, 90^{\circ}\right]$ and $\left[20^{\circ}, 30^{\circ}\right]$, respectively. We employ an SDF renderer~\cite{Jiang_2020_CVPR} since we represent the synthesized shapes with SDF grids.

% \noindent\textbf{Using Single-View Reconstruction Models.} It is straightforward that we can use a single-view reconstruction (SVR) model as the image-to-shape module in our framework. This way, we no longer need the shape prior network $G_M$ since the SVR model can map images to 3D shapes directly. A recent work named ISS~\cite{liu2022iss} also utilizes an SVR model to perform text-to-shape generation. However, ISS trains a mapper network to map CLIP features to the latent space of the SVR model which requires a two-stage fine-tuning to align the text and shape feature spaces. We show that this tedious fine-tuning process is redundant. With the help of an off-the-shelf text-to-image model, we can directly generate images from the text prompts and take the image as the input of the SVR model to synthesize shapes.

% \begin{figure}[t]
%     \centering
%     \includegraphics[width=\linewidth]{figures/svr.pdf}
%     \caption{Text-guided shape generation using our fine-tuned Stable Diffusion and DVR~\cite{Niemeyer_2020_CVPR}.}
%     \label{fig:svr}
% \end{figure}

\subsection{Fine-tuning Stable Diffusion for Rendering-Style Image Generation}
\label{sec:stylize}

\begin{figure}[t]
    \centering
    \includegraphics[width=1.0\linewidth]{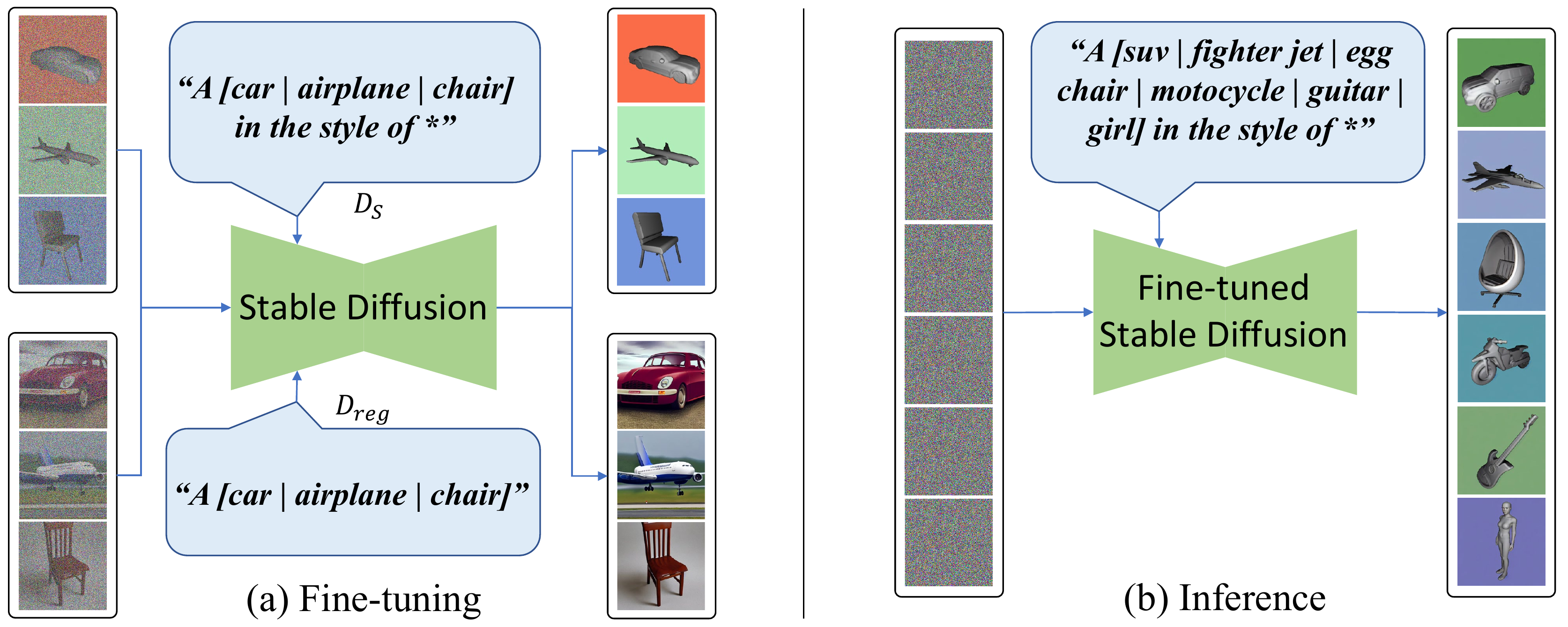}
    \caption{Fine-tuning Stable Diffusion into a stylized generator. (a) We fine-tune the text embedding $\boldsymbol{v}_{*}$ of a placeholder token $*$ and the weights of Stable Diffusion using a dataset $D_S$ that contains ShapeNet renderings and text descriptions in the format of \textit{``a CLS in the style of $*$"}. Additionally, we use another dataset $D_{reg}$, which contains images synthesized by the original Stable Diffusion with text prompts in the format of \textit{``a CLS"}, for regularization purposes. (b) With the fine-tuned Stable Diffusion, we can synthesize images that match the style of ShapeNet renderings by appending the postfix \textit{``in the style of $*$"} to the text prompt.}
    \label{fig:finetune}
    \vspace{-4mm}
\end{figure}

\begin{figure*}[t]
  \centering
   \includegraphics[width=\linewidth]{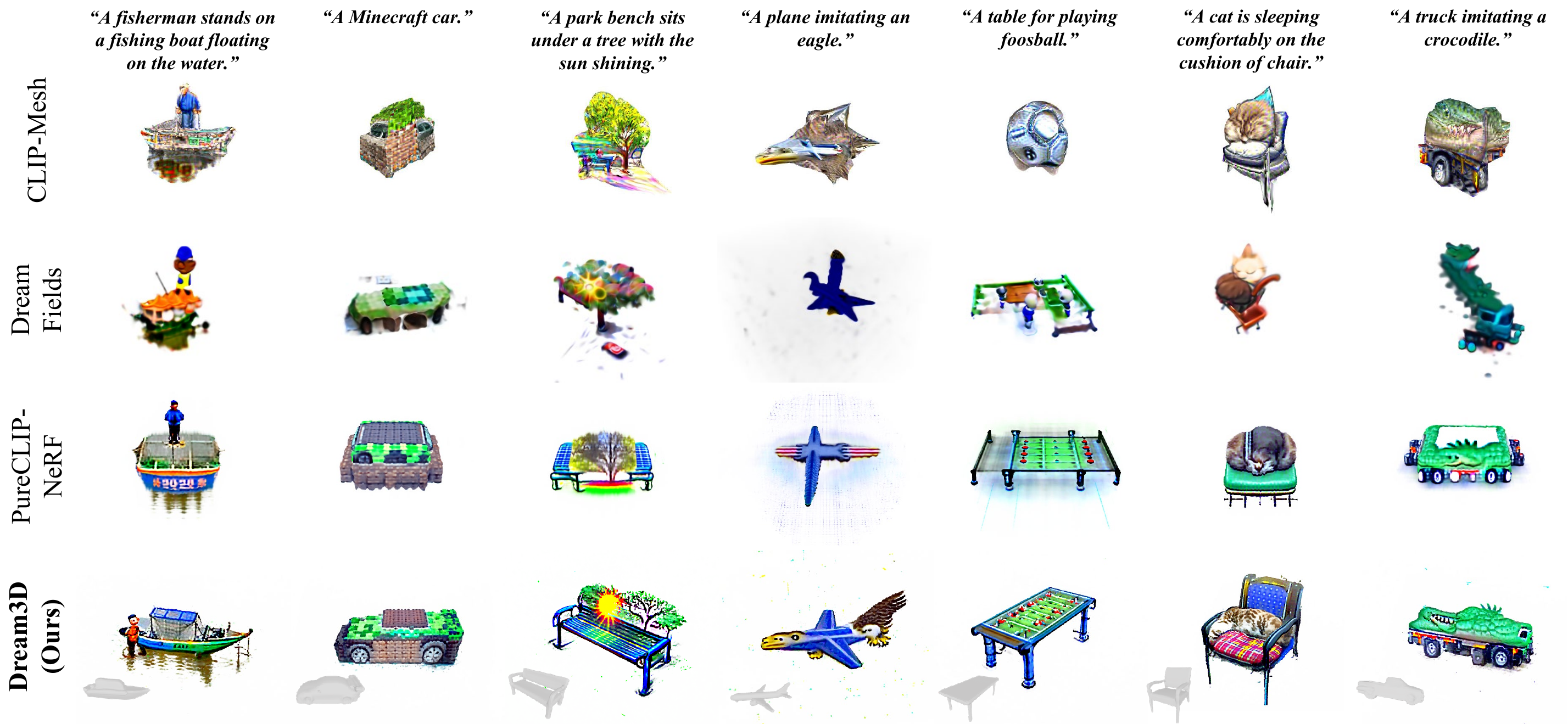}
   \vspace{-6mm}
   \caption{Qualitative comparison on text-guided 3D synthesis. The 3D shape prior used to initialize the CLIP-guided optimization process for each of our results (the last row) is also visualized below.}
   \label{fig:text23d}
   \vspace{-4mm}
\end{figure*}

As stated in \cref{sec:text2shape}, we use Stable Diffusion to directly bridge the text and image modalities for text-to-shape generation. Nonetheless, the image-to-shape module is trained on shape renderings, which exhibit a significant style domain gap from the images produced by Stable Diffusion. Previous research~\cite{Sanghi_2022_CVPR, liu2022iss} has attempted to combine a text-to-image model with an image-to-shape model for text-to-shape generation. However, this approach is plagued by the aforementioned style domain gap, leading to flawed geometric structures and diminished performance.

Inspired by recent work on controllable text-to-image generation such as textual inversion~\cite{gal2022image} and DreamBooth~\cite{ruiz2022dreambooth}, we propose a method for addressing the domain gap problem by fine-tuning Stable Diffusion into a stylized generator. Our core idea is to enable Stable Diffusion to replicate the style of shape renderings used to train the image-to-shape module outlined in \cref{sec:text2shape}. This allows us to seamlessly input the generated \textit{stylized} images into the image-to-shape module without being affected by the domain gap.

\noindent\textbf{Fine-tuning Process.} The fine-tuning process is illustrated in \cref{fig:finetune}. To fine-tune Stable Diffusion, we need a dataset that consists of shape renderings and related \textit{stylized} text prompts. For each shape $S$ in the ShapeNet dataset, we generate a set of shape renderings $\{\boldsymbol{I}_S^{j}\}_{j=1}^{N_S}$. Subsequently, each rendering $\boldsymbol{I}_S^{j}$ is linked with a \textit{stylized} text prompt $y_S^j$ in the format of \textit{``a CLS in the style of $*$"}, where \textit{CLS} denotes the shape category name and $*$ represents a placeholder token that requires optimization for its text embedding.
For instance, if the image is rendered from a chair shape, then the associated text prompt will be \textit{``a chair in the style of $*$"}. The paired dataset $D_S={(\boldsymbol{I}_S^{j}, y_S^j)}_{j=1}^{N_S}$ is then utilized to fine-tune Stable Diffusion by minimizing $\mathcal{L}_{\text{diffusion}}$ presented in \cref{eq:diffusion_loss_conditional}.

During fine-tuning, we freeze the CLIP text encoder of Stable Diffusion, and optimize two objectives: (i) the text embedding of the placeholder token $*$, denoted as $\boldsymbol{v}_*$, and (ii) the parameters $\theta$ of the diffusion model $\epsilon_{\theta}$. Optimizing the text embedding $\boldsymbol{v}_*$ aims to learn a virtual word that captures the style of the rendered images best, even though it is not present in the vocabulary of the text encoder. Fine-tuning the parameters $\theta$ of the diffusion model further enhances the ability to capture the style precisely since it is hard to control the synthesis of Stable Diffusion solely on the language level.  Our experiments demonstrate stable convergence of the fine-tuning process in approximately 2000 optimization steps, requiring only 40 minutes on a single Tesla A100 GPU. We show some synthesized results using the fine-tuned model in \cref{fig:finetune}.

%In case the fine-tuning process harms the generation capability of the original diffusion model and overfit on the synthetic dataset $D_S$, we adopt the prior-preserving loss introduced by DreamBooth~\cite{ruiz2022dreambooth}. Specifically, we synthesize a regularization dataset $D_{reg}=\{(\boldsymbol{I}_{reg}^{j}, y_{reg}^j)\}_{j=1}^{N_{reg}}$ before fine-tuning, where $y_{reg}^j$ denotes a text prompt in the format of "a CLS", and use both $D_S$ and $D_{reg}$ for fine-tuning. In our experiments, the fine-tuning process converges stably in around $2000$ optimization steps, which only takes about $40$ minutes on a single Tesla A100 GPU. We show some synthesized results using the fine-tuned model in \cref{fig:finetune}.

\noindent\textbf{Dataset Scale and Background Augmentation.}
We have identified two essential techniques empirically that enable the fine-tuned model to synthesize stylized images in a stable manner. Firstly, unlike textual inversion~\cite{gal2022image} or DreamBooth~\cite{ruiz2022dreambooth} which utilize only $3{-}5$ images, fine-tuning with a larger set of shape renderings containing thousands of images helps the model capture the style more precisely. Secondly, fine-tuning Stable Diffusion using shape renderings with a pure-white background results in a chaotic and uncontrollable background during inference. However, augmenting the shape renderings with random solid-color backgrounds allows the fine-tuned model to synthesize images with solid-color backgrounds stably, making it easy to remove the background if necessary. Further details can be found in the supplementary material.

% Denoting the fine-tuned Stable Diffusion model as $G_I$, we now obtain a \textit{stylized} text-to-image generator which can synthesize rendering-style images when conditioned on text prompts in the format of \textit{"a \{CLS\} in the style of $*$"}. By combing $G_I$ and the image-to-shape module, we bridge the text, image, and 3D shape modalities seamlessly and can synthesize high-quality 3D shapes from text prompts.

%------------------------------------------------------------------------
\section{Experiments}
\label{sec:experiments}
In this section, we evaluate the efficacy of our proposed text-to-3D synthesis framework. Initially, we compare our results with state-of-the-art techniques (\cref{sec:exp_text23d}). Subsequently, we demonstrate the effectiveness of our Stable-Diffusion-assisted approach for text-to-shape generation (\cref{sec:exp_text2shape}). Furthermore, we conduct ablation studies to assess the effectiveness of critical components of our framework (\cref{sec:exp_ablation}).

\noindent\textbf{Dataset.} 
Our framework requires only a set of untextured 3D shapes for training the 3D shape generator $G_S$. Specifically, we employ 13 categories from ShapeNet~\cite{chang2015shapenet} and utilize the data preprocessing procedure of Zheng \etal~\cite{zheng2022sdfstylegan} to generate $128^3$ SDF grids from the original meshes. During the fine-tuning of Stable Diffusion, we employ the SDF renderer~\cite{Jiang_2020_CVPR} to produce a shape rendering dataset.

\noindent\textbf{Implementation details.} 
The 3D generator $G_S$ utilizes the SDF-StyleGAN architecture. The diffusion-model-based shape embedding mapping network $G_M$ is based on an open-source DALL$\cdot$E 2 implementation\footnote{\url{https://github.com/lucidrains/DALLE2-pytorch}}. We train $G_M$ by extracting image embeddings from shape renderings using the CLIP ViT-B/32 image encoder. The stylized text-to-image generator $G_I$ is fine-tuned from Stable Diffusion v1.4. In the optimization stage, we set the learning rates for the density grid $\boldsymbol{V}_{\text{density}}$ and color MLP $f_{\text{rgb}}$ to $5\times 10^{-1}$ and $5\times 10^{-3}$ respectively, and we adopt the CLIP ViT-B/16 encoder as the guidance model. For each text prompt, we optimize for 5000 steps, while previous NeRF-based text-to-3D methods~\cite{Jain_2022_CVPR, lee2022understanding} typically require 10000 steps or more.

\noindent\textbf{Evaluation Metrics.} 
Regarding the primary results of our framework, \ie, text-guided 3D content synthesis, we report the CLIP retrieval precision on a manually created dataset of diverse text prompts and objects. For specifics regarding the dataset, please refer to the supplementary material. This metric quantifies the percentage of generated images that the CLIP encoder associates with the correct text prompt used for generation. We utilize Fréchet Inception Distance (FID)~\cite{heusel2017gans} to evaluate the shape generation quality for the initial text-to-shape generation stage.

\subsection{Text-Guided 3D Synthesis}
\label{sec:exp_text23d}
We compare our method with three state-of-the-art baseline methods on the task of text-guided 3D synthesis, \ie, DreamFields~\cite{Jain_2022_CVPR}, CLIP-Mesh~\cite{khalid2022clipmesh}, and PureCLIPNeRF~\cite{lee2022understanding}. We conduct tests using the default settings and official implementations for all baseline methods. In particular, we utilize the medium-quality configuration of DreamFields and the implicit architecture variant of PureCLIPNeRF owing to its superior performance.
% And the default optimization steps for DreamFields, CLIP-Mesh, and PureCLIPNeRF are $10000$, $5000$, and $15000$ respectively.

\begin{table}[t]
\footnotesize
\renewcommand{\arraystretch}{1.0}
\renewcommand{\tabcolsep}{4mm}
\centering
\begin{tabular}{lcc}
\toprule
\multirow{2}{*}{Method} & \multicolumn{2}{c}{CLIP R-Precision $\uparrow$} \\
                   & ViT-B/32       & ViT-B/16 \\ \hline
DreamFields~\cite{Jain_2022_CVPR}        &     63.24      &  92.65   \\
CLIP-Mesh~\cite{khalid2022clipmesh}     &  75.00      &      91.18   \\
PureCLIPNeRF~\cite{lee2022understanding}       &     73.53      &    88.24    \\
\hline
Ours w/o 3D prior  &     75.47      &        94.34        \\    
Ours               &     \textbf{85.29}      &      \textbf{98.53}          \\    
\bottomrule
\end{tabular}
\caption{Quantitative comparison on Text-guided 3D synthesis. All methods employ CLIP ViT/16 as the guiding model for optimization, while two distinct CLIP models are utilized to compute the CLIP retrieval precision.}
\label{tab:text23D}
\vspace{-4mm}
\end{table}

\noindent\textbf{Time cost.} 
Thanks to the shape prior initialization, our CLIP-guided optimization process exhibits significantly improved efficiency compared to previous NeRF-based methods. Dream3D optimizes for only 5000 steps within 25 minutes, while DreamFields and PureCLIPNeRF require more than 10000 steps, taking over an hour (measured on 1 A100 GPU). Training the 3D shape generator $G_S$ takes 7 days on 4 A100 GPUs and training the shape embedding generation network $G_M$ takes 1 day on 1 A100 GPU. It is worth noting that these models are only trained once and their inference time can be neglected compared to the optimization cost.

\begin{figure}[t]
  \centering
   \includegraphics[width=\linewidth]{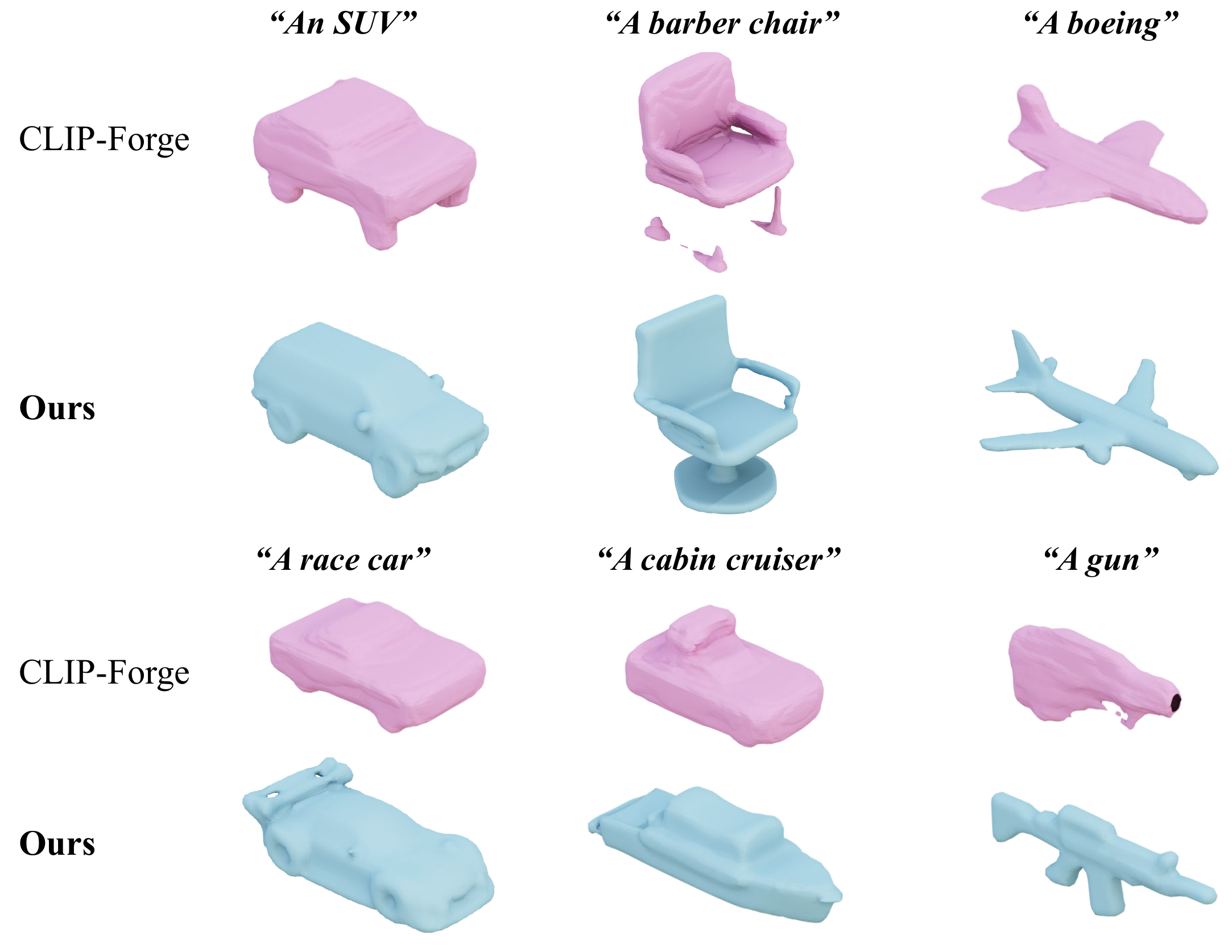}
   \vspace{-6mm}
   \caption{Text-guided 3D shape generation results. All the visualized meshes are extracted at the resolution of $64^3$. We can observe that our method can generate significantly more plausible 3D shapes benefitting from the high-quality 3D shape generator.}
   \label{fig:text2shape}
   \vspace{-2mm}
\end{figure}

\begin{table}[t]
\footnotesize
\renewcommand{\arraystretch}{1.0}
\renewcommand{\tabcolsep}{2mm}
\centering
\begin{tabular}{lccc}
\toprule
Method                                  & FID $\downarrow$  \\ 
\hline
CLIP-Forge~\cite{Sanghi_2022_CVPR}      &      112.38        \\
Ours w/o text-to-image                 &        58.36           \\  
Ours w/o fine-tuning SD                 &       61.88          \\    
Ours                                    &       \textbf{40.83}          \\    
\bottomrule
\end{tabular}
\vspace{-1mm}
\caption{Quantitative comparison on text-to-shape generation and ablation studies on the efficacy of fine-tuning Stable Diffusion.}
\label{tab:text2shape}
\vspace{-4mm}
\end{table}

\noindent\textbf{Quantitative Results.} 
We report the CLIP retrieval precision metrics in \cref{tab:text23D}. It is noteworthy that both the baseline methods and our approach utilize the CLIP ViT-B/16 encoder for optimization, and both the CLIP ViT-B/16 and CLIP ViT-B/32 encoders are employed as retrieval models. As Table \cref{tab:text23D} shows, our method achieves the highest CLIP R-Precision with both retrieval models. Moreover, our framework exhibits a significantly smaller performance gap between the two retrieval models compared to the baseline methods. By leveraging the 3D shape prior, our method initiates the optimization from a superior starting point, thereby mitigating the adversarial generation problem that prioritizes obtaining high CLIP scores while neglecting the visual quality. Consequently, our method demonstrates more robust performance across different CLIP models.

\noindent\textbf{Qualitative Results.}
The qualitative comparison is presented in \cref{fig:text23d}, indicating that the baseline methods~\cite{Jain_2022_CVPR, khalid2022clipmesh, lee2022understanding} encounter challenges in generating precise and realistic 3D objects, resulting in distorted and unrealistic visuals. DreamFields' results are frequently blurry and diffuse, while PureCLIPNeRF tends to synthesize symmetric objects. CLIP-Mesh experiences difficulties in generating intricate visual effects due to its explicit mesh representation. In contrast, our method effectively generates higher-quality 3D structures by incorporating explicit 3D shape priors.

\subsection{Text-to-Shape Generation}
\label{sec:exp_text2shape}

The research on zero-shot text-to-shape generation is limited, and we compare our approach with CLIP-Forge~\cite{Sanghi_2022_CVPR} and measure the quality of shape generation using the Fréchet Inception Distance (FID). Specifically, we synthesize 3 shapes for each prompt from a dataset of 233 text prompts provided by CLIP-Forge. Then, we render 5 images for each synthesized shape and compare them to a set of ground truth ShapeNet renderings to compute the FID. The ground truth images are obtained by randomly choosing 200 shapes from the test set of each ShapeNet category and rendering 5 views for each shape. 

As shown in \cref{tab:text2shape}, our approach achieves a lower FID than CLIP-Forge. CLIP-Forge employs a volumetric shape auto-encoder to generate 3D shapes. However, the qualitative results in \cref{fig:text2shape} indicate poor shape generation capability, making it difficult to generate plausible 3D shapes. A high-quality 3D shape prior is also advantageous for the optimization process as an excellent initialization.

\subsection{Ablation Studies}
\label{sec:exp_ablation}

\noindent\textbf{Effectiveness of Fine-tuning Stable Diffusion.} 
Our approach employs a fine-tuned Stable Diffusion to establish a connection between the text and image modalities. To evaluate its efficacy, we employ the original Stable Diffusion model to produce images from the text prompts of CLIP-Forge~\cite{Sanghi_2022_CVPR}. Subsequently, we employ these images to generate 3D shapes using the image-to-shape module. The results in the \nth{3} row of \cref{tab:text2shape} indicate a decline in FID performance, suggesting that this approach would negatively impact the shape generation process. Furthermore, we directly test using text embedding to generate shape embeddings with $G_M$, which also leads to a decline in performance as seen in the \nth{2} row of \cref{tab:text2shape}.

% \noindent\textbf{Random Color Augmentation for Fine-tuning.}

\noindent\textbf{Effectiveness of 3D Shape Prior.} 
To validate the efficacy of the 3D shape prior, we eliminate the first stage of our framework and optimize from scratch using the same text prompts as presented in \cref{sec:exp_text23d}. Subsequently, we evaluate the CLIP retrieval precision. The outcomes presented in \cref{tab:text23D} indicate that optimizing without 3D shape prior results in a considerable decline in performance, thereby demonstrating its effectiveness.

\begin{figure}[t]
    \centering
    \includegraphics[width=1.0\linewidth]{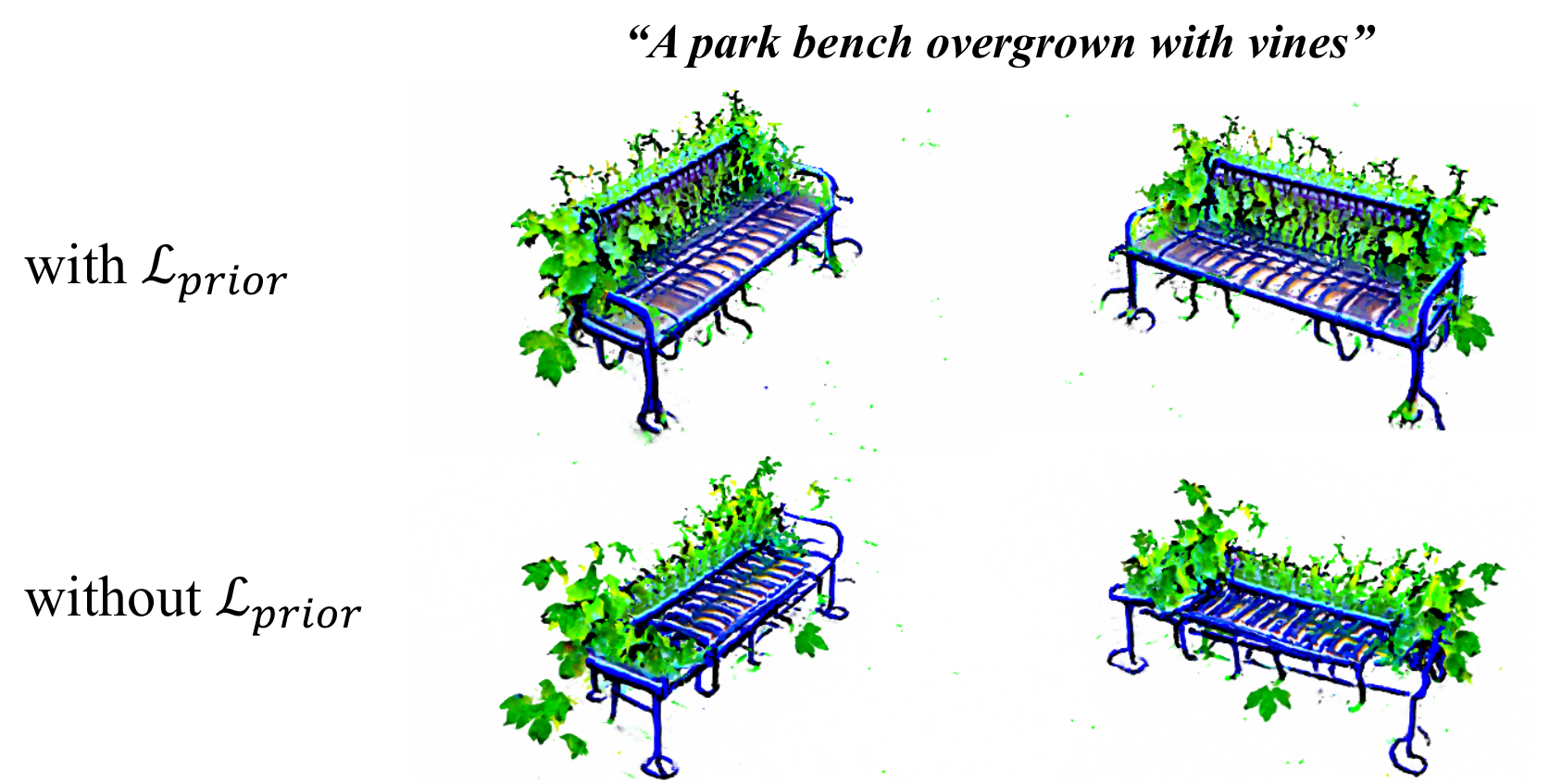}
    \caption{The effect of the 3D prior preserving loss $\mathcal{L}_{prior}$.}
    \label{fig:ablation}
    \vspace{-4mm}
\end{figure}

\noindent\textbf{Effectiveness of $\mathcal{L}_{\text{prior}}$.}
The 3D prior preserving loss $\mathcal{L}_{\text{prior}}$ shown in \cref{eq:prior_loss} aims to reinforce the 3D prior during the optimization process in case that the prior is gradually disturbed and discarded. To demonstrate its effectiveness, we synthesized \textit{``a park bench"} as a prior for the prompt \textit{``A park bench overgrown with vines"}, and then optimize with and without $\mathcal{L}_{\text{prior}}$ and compared the results, which are visualized in \cref{fig:ablation}. The results indicate that using $\mathcal{L}_{\text{prior}}$ during optimization helps maintain the structure of the 3D prior shape, while discarding it causes distortion and discontinuity artifacts, thereby disturbing the initial shape.

% \subsection{Optimization with Inaccurate Priors}

% \subsection{More Applications}
% \label{sec:exp_application}

% \noindent\textbf{Text-guided Shape Stylization.}

% \noindent\textbf{Combined 3D Content Creation.}

\section{Limitations and Future Work}
\label{sec:limitations}

Our framework relies on a fine-tuned Stable Diffusion to generate rendering-style images. Despite its strong generation capability, Stable Diffusion may produce shape images that fall outside the distribution of the training data of the image-to-shape module. This is due to the fact that Stable Diffusion is trained on an internet-scale text-image dataset, whereas the 3D shape generator is trained on ShapeNet.
Furthermore, the quality of text-to-shape synthesis in our framework is heavily reliant on the generation capability of the 3D generator. Our future work will explore incorporating stronger 3D priors into our framework to enable it to work with a wider range of object categories.

Additionally, our framework is indeed orthogonal to score distillation-based text-to-3D methods~\cite{poole2022dreamfusion, metzer2022latent, wang2022score}, as we can also utilize the score distillation sampling objective for optimization. We believe that incorporating 3D shape priors can enhance the quality and diversity of the generation results, as DreamFusion~\cite{poole2022dreamfusion} acknowledged.

\section{Conclusion}
\label{sec:conclusion}

This paper introduces Dream3D, a text-to-3D synthesis framework that can generate diverse and imaginative 3D content from text prompts. Our approach incorporates explicit 3D shape priors into the CLIP-guided optimization process to generate more plausible 3D structures. To address the text-to-shape generation, we propose a straightforward yet effective method that utilizes a fine-tuned text-to-image diffusion model to bridge the text and image modalities. Our method is shown to generate 3D content with superior visual quality and shape accuracy compared to previous work, as demonstrated by extensive experiments.

\noindent\textbf{Acknowledgements.} 
The work was supported by National Key R\&D Program of China (2018AAA0100704), NSFC \#61932020, \#62172279, Science and Technology Commission of Shanghai Municipality (Grant No. 20ZR1436000), Program of Shanghai Academic Research Leader, and ``Shuguang Program" supported by Shanghai Education Development Foundation and Shanghai Municipal Education Commission.

% \begin{figure}[t]
%   \centering
%   \fbox{\rule{0pt}{2in} \rule{0.9\linewidth}{0pt}}
%    %\includegraphics[width=0.8\linewidth]{egfigure.eps}

%    \caption{Example of caption.
%    It is set in Roman so that mathematics (always set in Roman: $B \sin A = A \sin B$) may be included without an ugly clash.}
%    \label{fig:onecol}
% \end{figure}

% \begin{figure*}
%   \centering
%   \begin{subfigure}{0.68\linewidth}
%     \fbox{\rule{0pt}{2in} \rule{.9\linewidth}{0pt}}
%     \caption{An example of a subfigure.}
%     \label{fig:short-a}
%   \end{subfigure}
%   \hfill
%   \begin{subfigure}{0.28\linewidth}
%     \fbox{\rule{0pt}{2in} \rule{.9\linewidth}{0pt}}
%     \caption{Another example of a subfigure.}
%     \label{fig:short-b}
%   \end{subfigure}
%   \caption{Example of a short caption, which should be centered.}
%   \label{fig:short}
% \end{figure*}

% \begin{table}
%   \centering
%   \begin{tabular}{@{}lc@{}}
%     \toprule
%     Method & Frobnability \\
%     \midrule
%     Theirs & Frumpy \\
%     Yours & Frobbly \\
%     Ours & Makes one's heart Frob\\
%     \bottomrule
%   \end{tabular}
%   \caption{Results.   Ours is better.}
%   \label{tab:example}
% \end{table}

%%%%%%%%% REFERENCES
{\small
\bibliographystyle{ieee_fullname}
\bibliography{reference}
}

\clearpage
\clearpage
\newpage

\renewcommand\thesection{\Alph{section}}
\setcounter{section}{0}

\section{Details of 3D Generator $G_S$}
\label{sec_supp:generator}
\begin{figure*}[t]
  \centering
   \includegraphics[width=\linewidth]{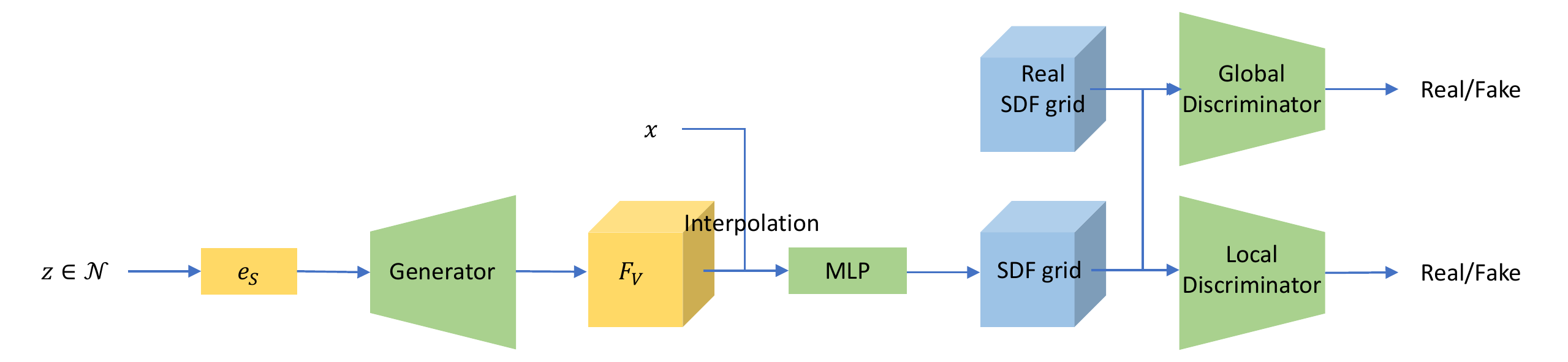}
   \caption{The network architecture of the 3D generator based on SDF-StyleGAN~\cite{zheng2022sdfstylegan}.}
   \label{fig_supp:generator}
\end{figure*}

We adopt the architecture of SDF-StyleGAN~\cite{zheng2022sdfstylegan} as our 3D generator. As \cref{fig_supp:generator} shows, it maps a random noise $\boldsymbol{z} \sim \mathcal{N}(\boldsymbol{0}, \boldsymbol{I})$ to a latent shape embedding $\boldsymbol{e}_S\in \mathcal{W}$ and synthesizes a 3D feature volume $\boldsymbol{F}_V$, which is an implicit representation of the generated shape. 
We can query the SDF value at arbitrary position $\boldsymbol{x}$ by feeding the interpolated feature from $\boldsymbol{F}_V$ at $\boldsymbol{x}$ into a jointly trained MLP network. 
During training, a global discriminator and a local discriminator are used simultaneously to supervise the generated SDF grids at the coarse and fine level respectively. Different from the original SDF-StyleGAN that trains one network for \textit{one shape category}, we train \textit{one} 3D shape generator $G_S$ on \textit{13 categories} of the ShapeNet~\cite{chang2015shapenet} dataset to enlarge the shape generation capability.

\section{Details of Shape Embedding Mapping Network $G_M$}
\label{sec_supp:diffusion}

The shape embedding mapping network $G_M$ is a diffusion-model-based generative network that can generate shape embeddings $\boldsymbol{e}_S$ from the CLIP image embeddings $\boldsymbol{e}_I$ of shape renderings. 
The network architecture and training strategy of $G_M$ are based on an open-source DALL-E-2~\cite{ramesh2022hierarchical} implementation\footnote{\url{https://github.com/lucidrains/DALLE2-pytorch}}. 
Specifically, $G_M$ is equivalent to the \textit{diffusion prior network} in DALL-E-2 which generates CLIP image embeddings from CLIP text embeddings. Here we replace the input with CLIP image embeddings of shape renderings and the output with shape embeddings.
We use the \textit{DiffusionPrior} class in the codebase to implement $G_M$ and the \textit{train\_diffusion\_prior.py} script to train $G_M$. The model and training hyperparameters are listed in \cref{tab_supp:diffusion_params}.

\begin{figure}
  \centering
  \begin{subfigure}{\linewidth}
    \includegraphics[width=\linewidth]{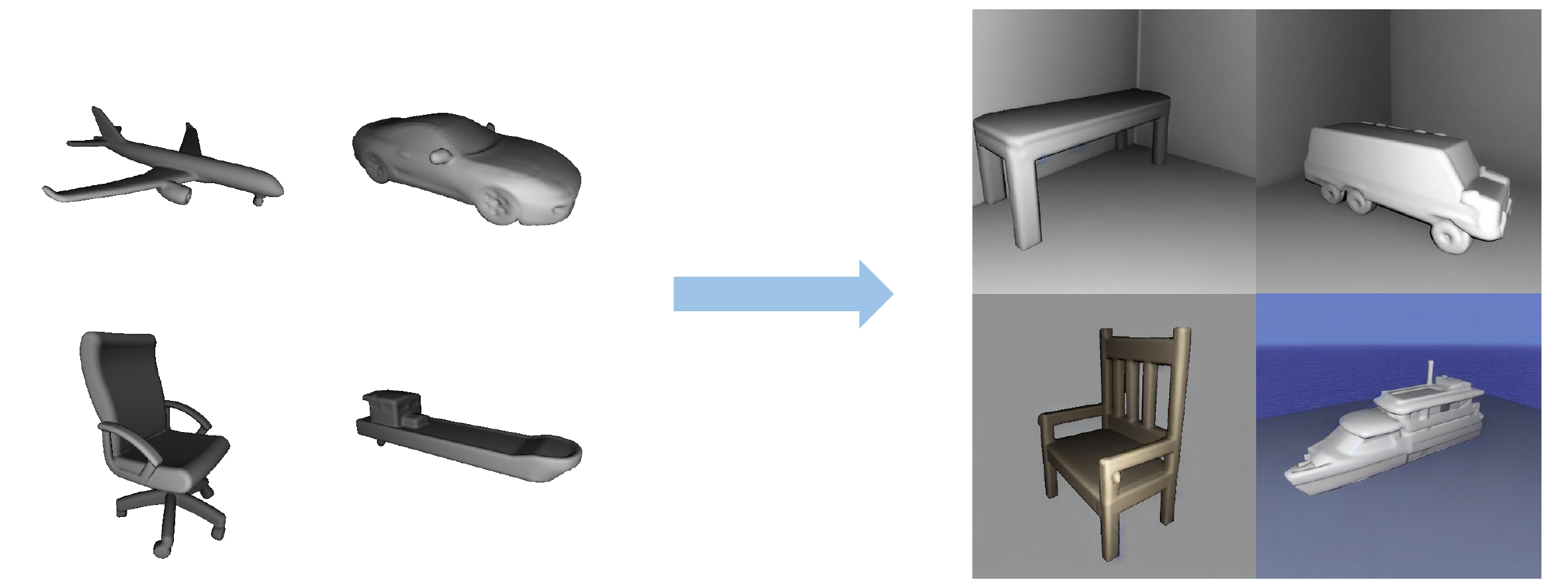}
    \caption{Solid white background.}
    \label{fig_supp:finetune_white}
  \end{subfigure}
  \begin{subfigure}{\linewidth}
    \includegraphics[width=\linewidth]{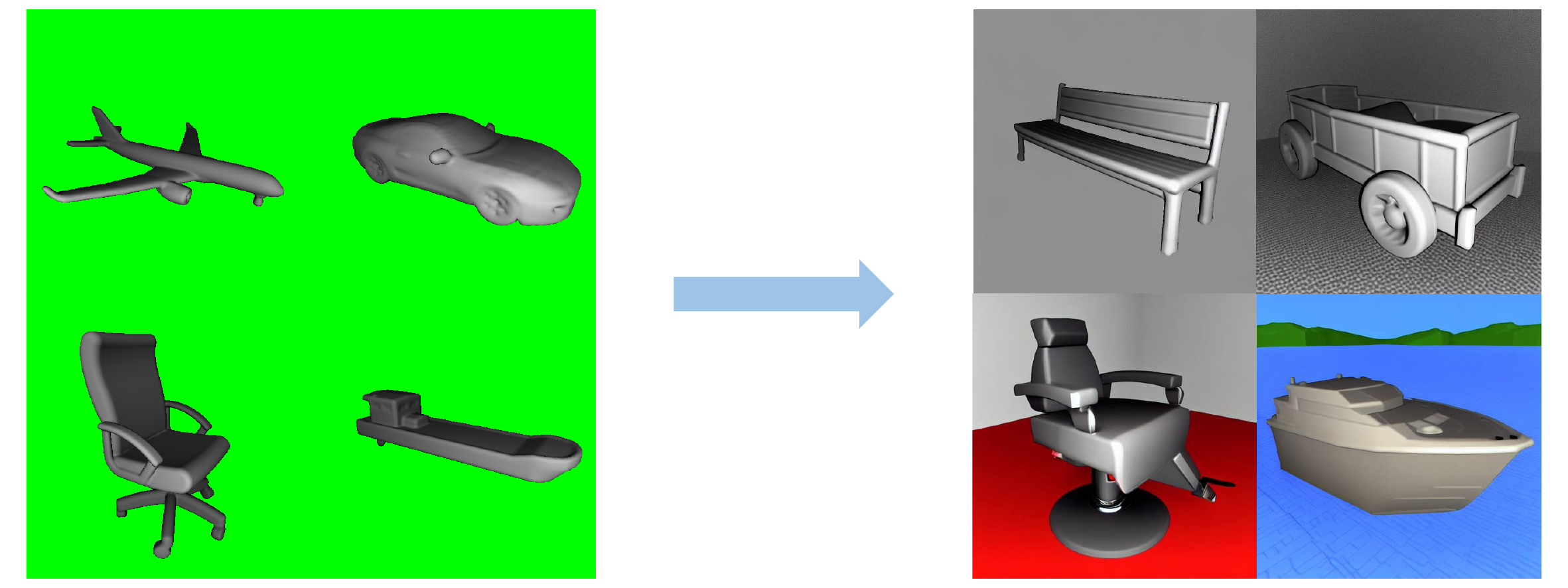}
    \caption{Solid green background.}
    \label{fig_supp:finetune_green}
  \end{subfigure}
  \begin{subfigure}{\linewidth}
    \includegraphics[width=\linewidth]{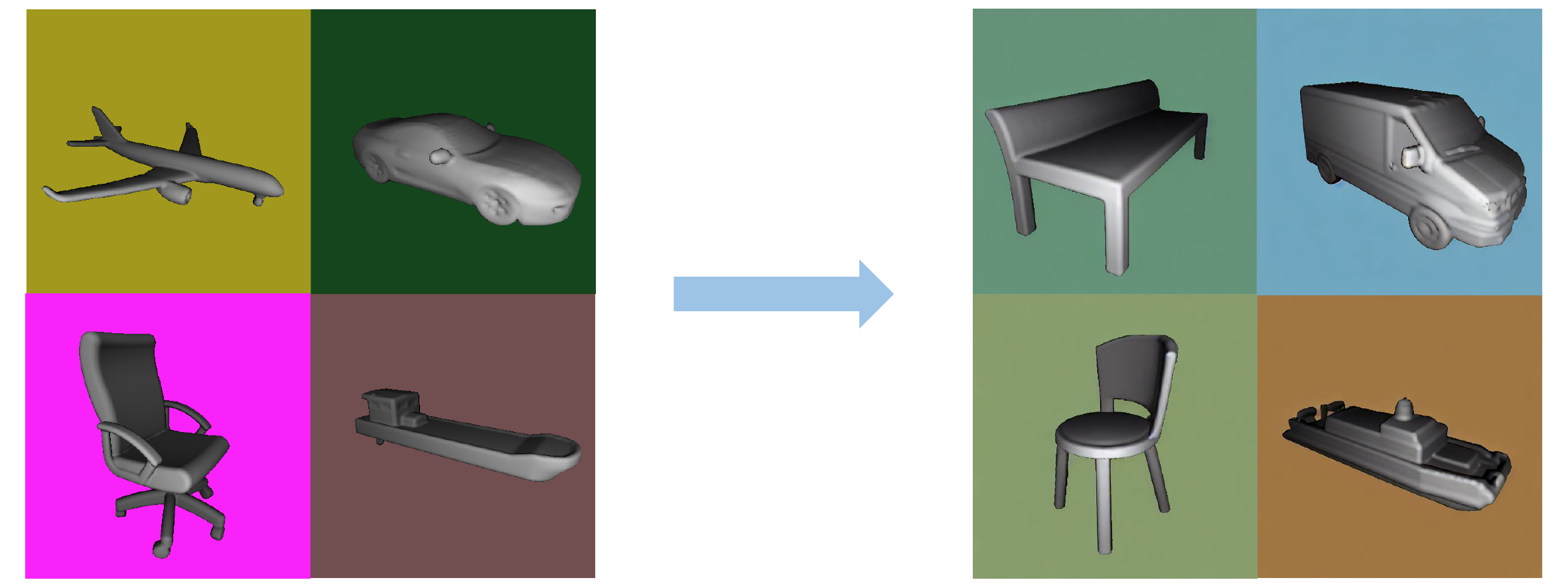}
    \caption{Random-color background.}
    \label{fig_supp:finetune_bg}
  \end{subfigure}
  \caption{The results of fine-tuning Stable Diffusion using shape renderings with different backgrounds. For each type of background, we visualize the shape renderings used to fine-tune Stable Diffusion on the left and the images synthesized by the fine-tuned Stable Diffusion on the right.}
  \label{fig_supp:finetune}
\end{figure}

\section{Details of Fine-tuning Stable Diffusion}
\label{sec_supp:finetune}
In our framework, we connect the text and image modalities by fine-tuning the Stable Diffusion into a stylized generator with a set of shape renderings $\{\boldsymbol{I}_S^{j}\}_{j=1}^{N_S}$ and give it the ability to synthesize images in the ``rendering" style. In experiments, we find it crucial to utilize a large set of shape renderings for fine-tuning and to augment the backgrounds of the shape renderings with random colors. 

We tried fine-tuning Stable Diffusion using shape renderings with three different types of backgrounds: 1) solid white background, 2) solid green background and 3) random-color background. We visualize the shape renderings used for fine-tuning and the images synthesized by the fine-tuned Stable Diffusion in \cref{fig_supp:finetune}. As \cref{fig_supp:finetune_white} and \cref{fig_supp:finetune_green} show, although shape renderings with solid-white or solid-green backgrounds can make the fine-tuned Stable Diffusion capture the ``rendering" style of the object successfully, the backgrounds in the synthesized images are out of control, \ie, the fine-tuned Stable Diffusion fails to synthesize images with solid-color backgrounds. This will increase the difficulty of separating the foreground objects from the backgrounds and affect the stability of the subsequent image-to-shape generation since the shape embedding mapping network $G_M$ is trained on shape renderings with solid-color backgrounds. In comparison, augmenting the backgrounds of the shape renderings with random colors leads to a stable stylized generator that can synthesize solid-color-background images consistently, as \cref{fig_supp:finetune_bg} shows. 

During fine-tuning, we indeed expect the Stable Diffusion model to capture two types of styles: 1) the ``rendering" style of the foreground object and 2) the ``solid-color" style of the background. Similar to the observation that the foreground ``rendering" style requires a large set of rendered images to learn, we consider that a single-color background is too few to be recognized as a ``solid-color background style" by the Stable Diffusion model, while showing a lot of different solid-color examples to the model can make it notice the solid-color background style and capture it during fine-tuning.

\begin{table}[t]
\footnotesize
\renewcommand{\arraystretch}{1.0}
\renewcommand{\tabcolsep}{2mm}
\centering
\begin{tabular}{lc|lc}
\toprule
Model Parameter & Value & Training Parameter & Value \\
\hline
timesteps  & 100 & iterations & 500,000  \\
beta\_schedule & cosine & max\_grad\_norm & 0.5 \\
predict\_x\_start & True & batch\_size & 1024 \\
cond\_drop\_prob & 0 & learning\_rate & $1.1\times 10^{-4}$ \\
dim & 512 & weight\_decay & $6.02\times 10^{-2}$ \\
depth & 6 & ema\_beta & 0.9999 \\
dim\_head & 64 & ema\_update\_every & 10 \\
heads & 8 & Adam $\beta_1, \beta_2$ & 0.9, 0.999 \\
\bottomrule
\end{tabular}
\caption{Model details and training hyper-parameters of the shape embedding mapping network $G_M$.}
\label{tab_supp:diffusion_params}
\end{table}

\begin{table}[t]
\footnotesize
\renewcommand{\arraystretch}{1.0}
\renewcommand{\tabcolsep}{4mm}
\centering
\begin{tabular}{lc}
\toprule
Background & FID $\downarrow$ \\
\hline
Solid white & 60.61 \\
Solid green & 71.04 \\
Random-color & \textbf{33.71} \\
\bottomrule
\end{tabular}
\caption{The Fréchet Inception Distance (FID) between the shape renderings used for fine-tuning and the images synthesized by the fine-tuned Stable Diffusion.}
\label{tab_supp:finetune_fid}
\end{table}

To better demonstrate the importance of the random-color background augmentation, we also evaluate the Fréchet Inception Distance (FID) between the shape renderings used for fine-tuning and the images synthesized by the fine-tuned Stable Diffusion in \cref{tab_supp:finetune_fid}. For each type of background, we render $1000$ images with that background for each ShapeNet~\cite{chang2015shapenet} category, forming a shape rendering dataset containing 13000 images in total (denoting as $D_S$). Then we leverage $D_S$ to fine-tune the Stable Diffusion model for 5000 steps, and utilize the fine-tuned Stable Diffusion to synthesize 100 images for each shape category using the text prompt \textit{"a CLS in the style of *"}, leading to a set of 1300 generated images (denoting as $D_{gen}$). Finally, we compute the FID between $D_S$ and $D_{gen}$. As \cref{tab_supp:finetune_fid} shows, augmenting the backgrounds of shape renderings with random colors significantly boosts the FID, which demonstrates its effectiveness.

\section{Details of 3D Optimization with 3D Shape Prior}
\label{optimization}

\subsection{DVGO-based Volume Rendering}
\label{sec_supp:dvgo}
In the optimization stage, we adopt DVGO~\cite{Sun_2022_CVPR_DVGO} as our 3D scene representation which represents NeRF~\cite{mildenhall2020nerf} with a density voxel grid $\boldsymbol{V}_{density} \in \mathbb{R}^{N_x\times N_y\times N_z}$ and a shallow color MLP network $f_{rgb}$ for efficient optimization. Given a 3D position $\boldsymbol{x}$, we query its density $\sigma$ and color $c$ by:
\begin{subequations}
\begin{align}
    \tilde{\sigma} &=\operatorname{interp}(\boldsymbol{x}, \boldsymbol{V}_{density}), \label{eq_supp:dvgo_query1} \\
    \sigma &=\operatorname{softplus}(\tilde{\sigma}) = \log(1+\exp(\tilde{\sigma} + b)), \label{eq_supp:dvgo_query2} \\
    c &=f_{rgb}(\gamma(\boldsymbol{x})), \label{eq_supp:dvgo_query3}
\end{align}
\label{eq_supp:dvgo_query}
\end{subequations}
where $\gamma(\cdot)$ denotes a positional encoding function. $\operatorname{interp}(\cdot)$ denotes the trilinear interpolation. The shifted softplus function $\operatorname{softplus}(\cdot)$ is applied to transform the raw density value $\tilde{\sigma}$ into activated density value $\sigma$ (\ie, a mapping of $\mathbb{R} \rightarrow \mathbb{R}_{\geq 0}$), the shift $b$ is a hyperparameter. To be noted, the density grid $\boldsymbol{V}_{density}$ stores the raw density values instead of the activated ones. DVGO~\cite{Sun_2022_CVPR_DVGO} calls the scheme of interpolating on the raw density values first and then performing softplus activation as "post-activation" and demonstrates its advantages on producing sharper shape boundaries over other choices.

To render the color of a pixel $\hat{C}(r)$, we cast the ray $r$ from the camera center through the pixel, and sample $K$ points between the pre-defined near and far planes. We then query the densities and colors of the $K$ ordered sampled points $\left\{\left(\sigma_i, c_i\right)\right\}_{i=1}^K$ using \cref{eq_supp:dvgo_query}. Finally, we accumulate the $K$ queried results into a single color with the volume rendering process:
\begin{subequations}
\begin{align}
    \hat{C}(r) &=\left(\sum_{i=1}^K T_i \alpha_i c_i\right)+T_{K+1} c_{bg}, \label{eq_supp:dvgo_rendering1} \\
    \alpha_i &=\operatorname{alpha}\left(\sigma_i, \delta_i\right)=1-\exp \left(-\sigma_i \delta_i\right), \label{eq_supp:dvgo_rendering2} \\
    T_i &=\prod_{j=1}^{i-1}\left(1-\alpha_j\right), \label{eq_supp:dvgo_rendering3}
\end{align}
\end{subequations}
where $\alpha_i$ denotes the opacity representing the probability of termination at point $i$, $T_i$ denotes the accumulated transmittance from the near plane to point $i$, $\delta_i$ denotes the distance to the adjacent sampled point, and $c_{bg}$ demotes a pre-defined background color.

Following DVGO, all values in $\boldsymbol{V}_{density}$ are initialized as $0$ and the bias term in \cref{eq_supp:dvgo_query2} is set to
\begin{equation}
b=\log \left(\left(1-\alpha_{init}\right)^{-\frac{1}{s}}-1\right),
\label{eq_supp:dvgo_init}
\end{equation}
where $\alpha_{init}$ is a hyperparameter and is set to $10^{-6}$ in practice. With such an initialization, the accumulated transmittance $T_i$ is decayed by $1-\alpha_{init} \approx 1$ for a ray that traces forward a distance of a voxel size $s$, making the scene ``transparent" at the beginning of optimization.

\begin{figure}[t]
  \centering
   \includegraphics[width=\linewidth]{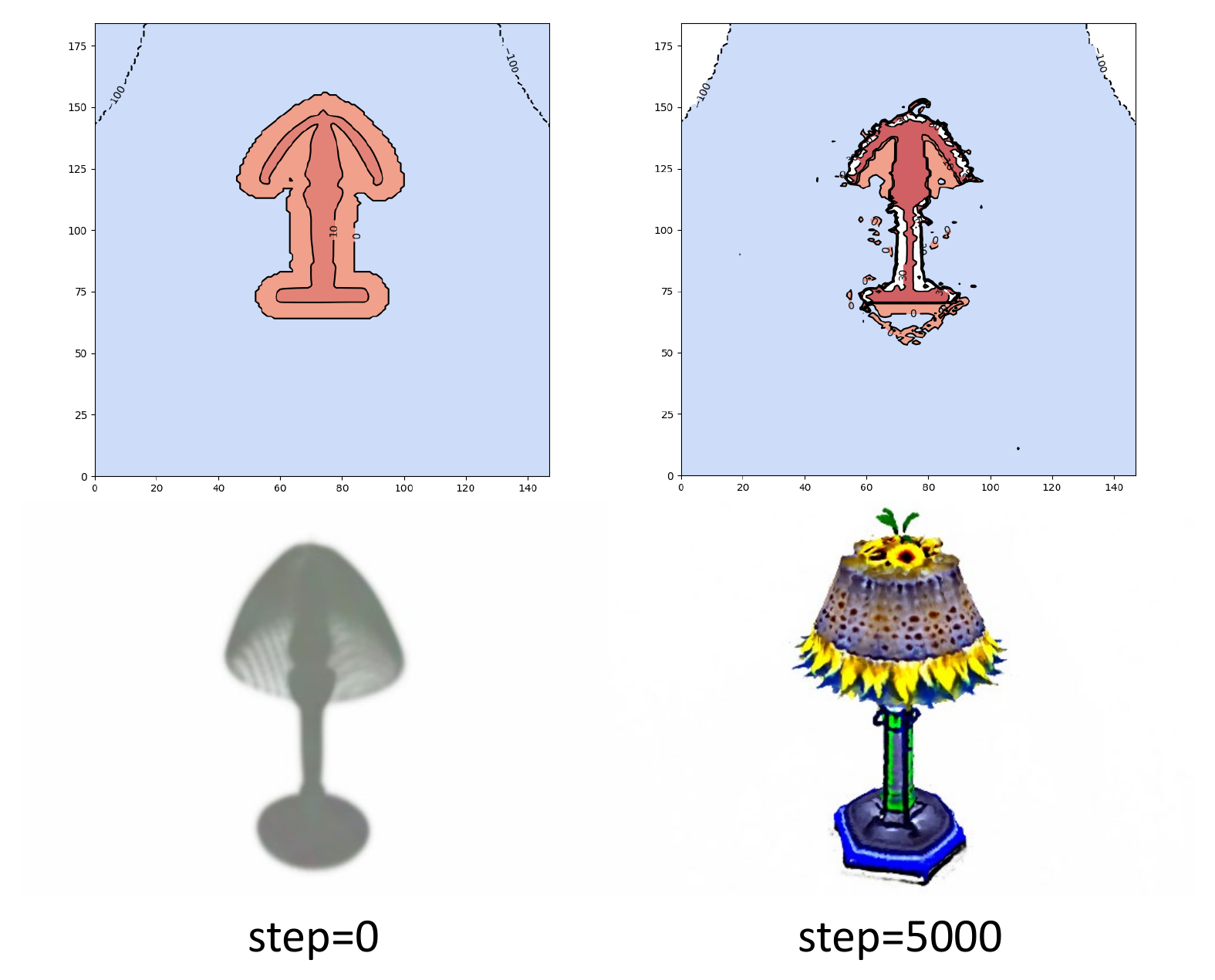}
   \caption{Illustration on 3D optimization with 3D shape prior. The text prompt is \textit{"A lamp imitating sunflower"}. We visualize the contours of the density grid (up) and the volume-rendered images (bottom) at the $0^{th}$ and $5000^{th}$ optimization steps to show how the density grid is initialized and optimized.}
   \label{fig_supp:optimize_prior}
\end{figure}

\subsection{Shape Prior Initialization and Optimization}
A big difference between our text-guided 3D synthesis framework and previous methods~\cite{Jain_2022_CVPR, khalid2022clipmesh, lee2022understanding} is that we use an explicit \textit{3D shape prior} to initialize the CLIP-guided optimization process, instead of optimizing from a \textit{randomly-initialized} 3D representation. Given a 3D shape prior $S$ represented by an SDF grid $\tilde{\boldsymbol{V}}_{sdf} \in \mathbb{R}^{N_x\times N_y\times N_z}$, we use it to initialize the density voxel grid $\boldsymbol{V}_{density}$ with the following equations~\cite{yariv2021volume, or2022stylesdf, Sun_2022_CVPR_DVGO}:
\begin{subequations}
\begin{align}
    \boldsymbol{\Sigma} &= \frac{1}{\beta}\operatorname{sigmoid}\left(-\frac{\tilde{\boldsymbol{V}}_{sdf}}{\beta}\right), \label{eq_supp:sdf2density1} \\
    \boldsymbol{V}_{density} &= \max(0, \operatorname{softplus}^{-1}(\boldsymbol{\Sigma})), \label{eq_supp:sdf2density2}
\end{align}
\end{subequations}
where $\operatorname{sigmoid}(x)=1/(1+e^{-x})$ and $\operatorname{softplus}^{-1}(x)=\log(e^x-1)$. \cref{eq_supp:sdf2density1} converts SDF values to activated density values (equivalent to the $\sigma$ in \cref{eq_supp:dvgo_query2}), where $\beta>0$ controls the sharpness of the shape boundary, and smaller $\beta$ leads to a sharper shape boundary. We set $\beta=0.05$ in our experiments. \cref{eq_supp:sdf2density2} further transforms the activated density values into raw density values (equivalent to the $\tilde{\sigma}$ in \cref{eq_supp:dvgo_query1}). 

With such an initialization, the density values on the shape surface will be close to $\frac{1}{2\beta}$ ($\log(\exp(\frac{1}{\beta}\cdot\operatorname{sigmoid}(0))-1)\approx \frac{1}{2\beta}$). The area inside the shape surface will have larger density values ($>\frac{1}{2\beta}$), and the density values outside the shape will decrease with the distance from the shape surface. We set the minimum density value outside the shape to $0$ so the area far from the shape surface has the same initialization as the original DVGO. 

As \cref{fig_supp:optimize_prior} shows, at the beginning of the 3D optimization process (step=0), the 3D shape prior is visible due to the larger density values around the shape surface. As a result, the area around the shape surface will dominate the volume rendering, and the density/color values in this area will be updated faster than the area far from the surface. Based on the initialization, Then subsequent CLIP-guided optimization process further provides more flexibility and is able to synthesize more diverse structures and textures.

\section{Additional Results on Text-to-Shape Generation}
\label{sec_supp:text2shape}
\begin{figure}[t]
  \centering
   \includegraphics[width=\linewidth]{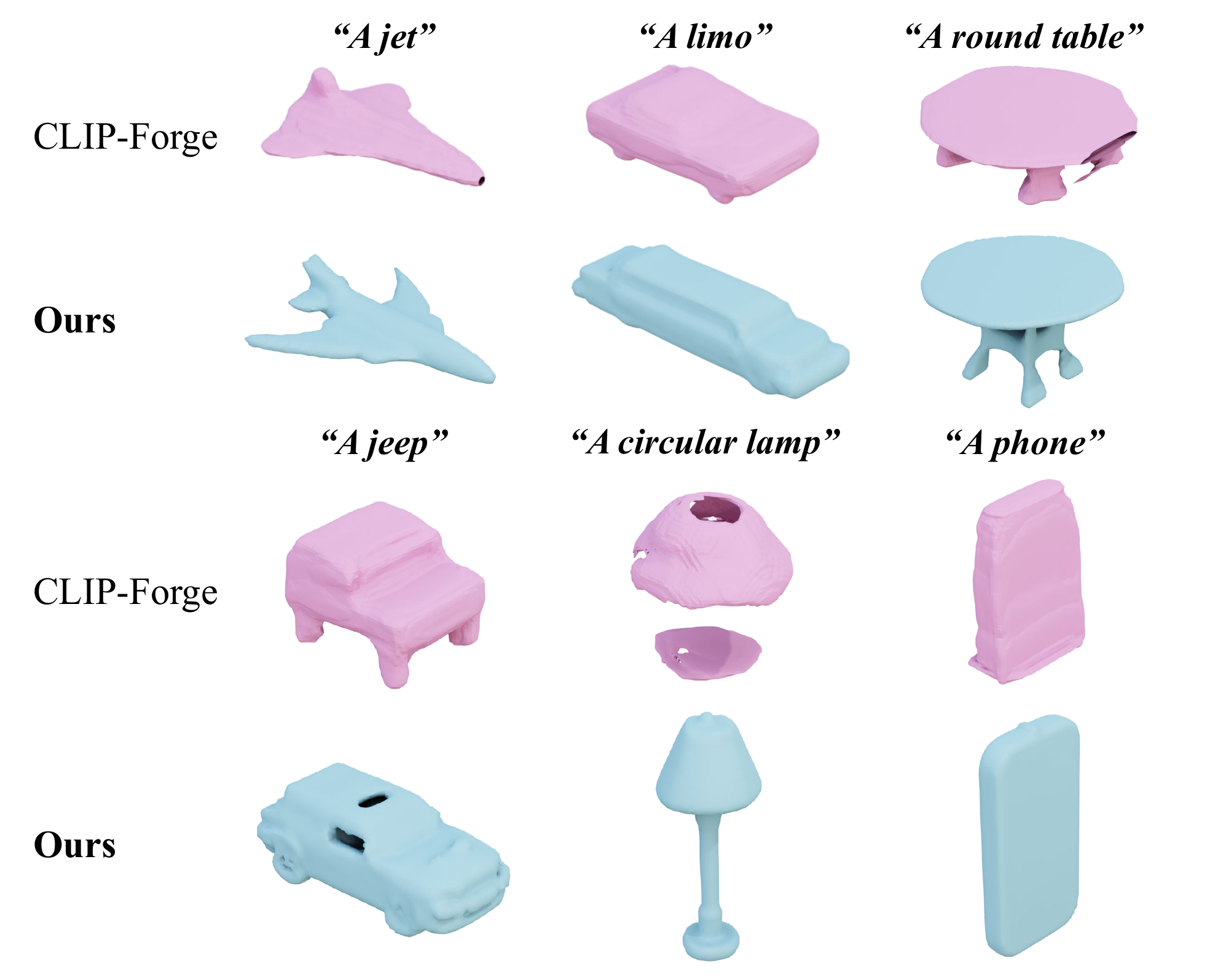}
   \caption{Additional text-guided shape generation results. All meshes are extracted at $64^3$ resolution.}
   \label{fig_supp:text2shape}
\end{figure}

We show additional qualitative text-guided 3D shape generation results in \cref{fig_supp:text2shape}. Compared to CLIP-Forge~\cite{Sanghi_2022_CVPR}, our method produces more plausible 3D shapes thanks to the high-quality 3D generator, while the shapes generated by ~\cite{Sanghi_2022_CVPR} suffer from rough surfaces and discontinuities.

\begin{table}[t]
\footnotesize
\renewcommand{\arraystretch}{1.0}
\renewcommand{\tabcolsep}{4mm}
\centering
\begin{tabular}{lccc}
\toprule
Method & FID $\downarrow$ & FPD $\downarrow$ & MMD $\uparrow$ \\
\hline
CLIP-Forge~\cite{Sanghi_2022_CVPR} & 112.38 & 6.896  & 0.670  \\
Ours & \textbf{40.83} & \textbf{1.301}  & \textbf{0.725}  \\
\bottomrule
\end{tabular}
\caption{Additional quantitative results compared with CLIP-Forge on text-guided shape generation.}
\label{tab_supp:text2shape}
\end{table}

Besides, we also provide more quantitative comparisons with CLIP-Forge on text-to-shape generation. We generate 3 shapes for each text prompt in the text prompt set provided by CLIP-Forge and measure three metrics: 1)  Fréchet Inception Distance (FID)~\cite{heusel2017gans} between 5 rendered images for each shape with different camera poses and a set of images rendered from the ground truth shapes in the ShapeNet dataset with the same camera poses. 2) Fréchet Point Distance (FPD)~\cite{Shu_2019_ICCV}, for each generated shape and each ground truth shape in the ShapeNet test set, we extract the mesh at $64^3$ resolution and sample 2048 points from the mesh surface, then pass the points to a DGCNN~\cite{wang2019dynamic} backbone network pre-trained on the point cloud classification task and use the feature of the last layer to compute this metric. 3) Maximum Measure Distance (MMD), for each generated shape represented by a $32^3$ occupancy grid, we match a shape in the ShapeNet test set based on the highest IOU, and then average the IOU across all the text queries. As \cref{tab_supp:text2shape} shows, our text-to-shape generation method outperforms CLIP-Forge on all three metrics.

\section{Additional Results on Text-to-3D Synthesis}
\label{sec_supp:text23d}

\begin{figure*}[th]
  \centering
   \includegraphics[width=\linewidth]{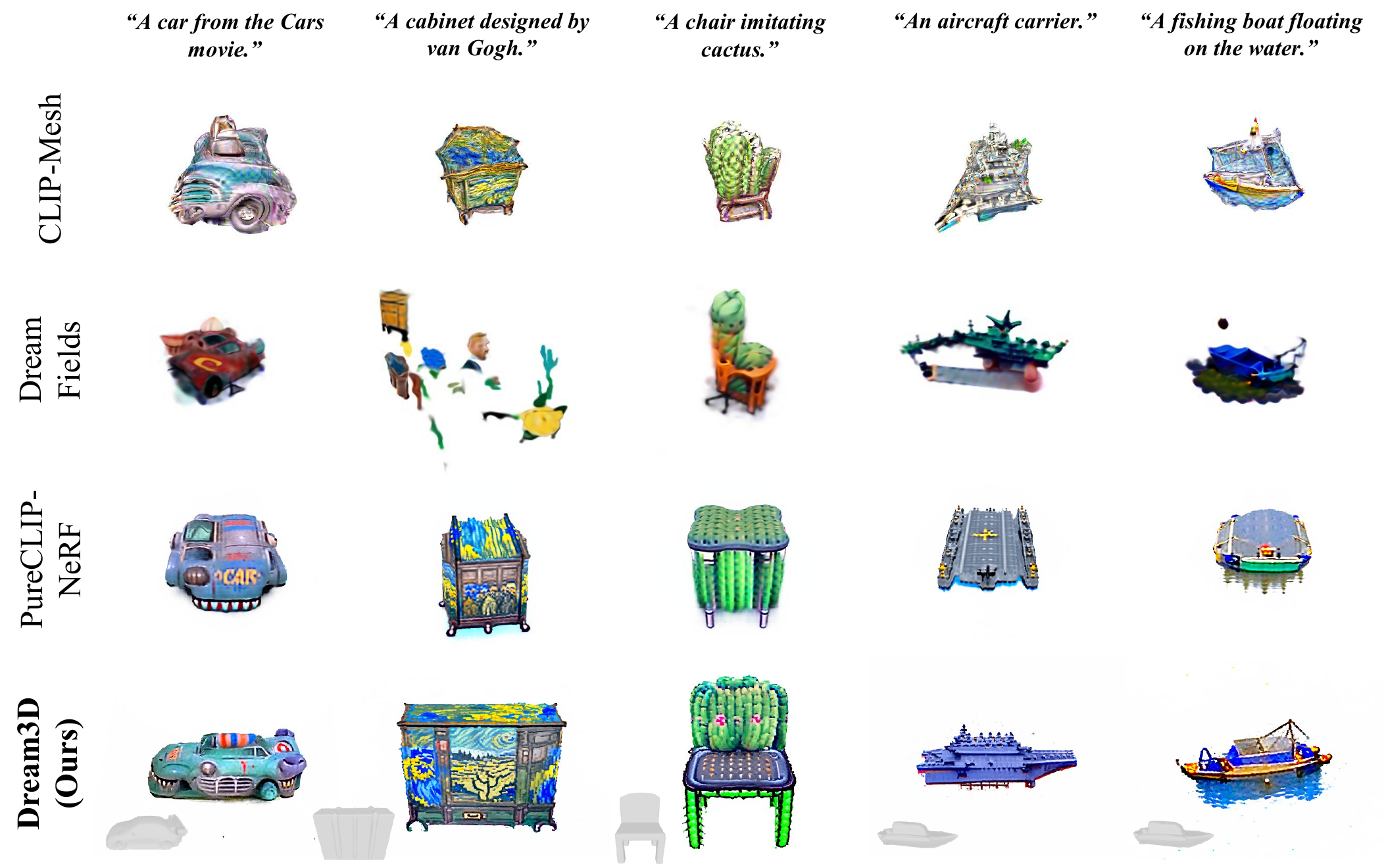}
   \caption{Additional qualitative comparisons on text-guided 3D synthesis. For each of our results (the last row), we also visualize the 3D shape prior
used to initialize the CLIP-guided optimization process below it.}
   \label{fig_supp:text23d_comparison}
\end{figure*}

\begin{figure*}[th]
  \centering
   \includegraphics[width=\linewidth]{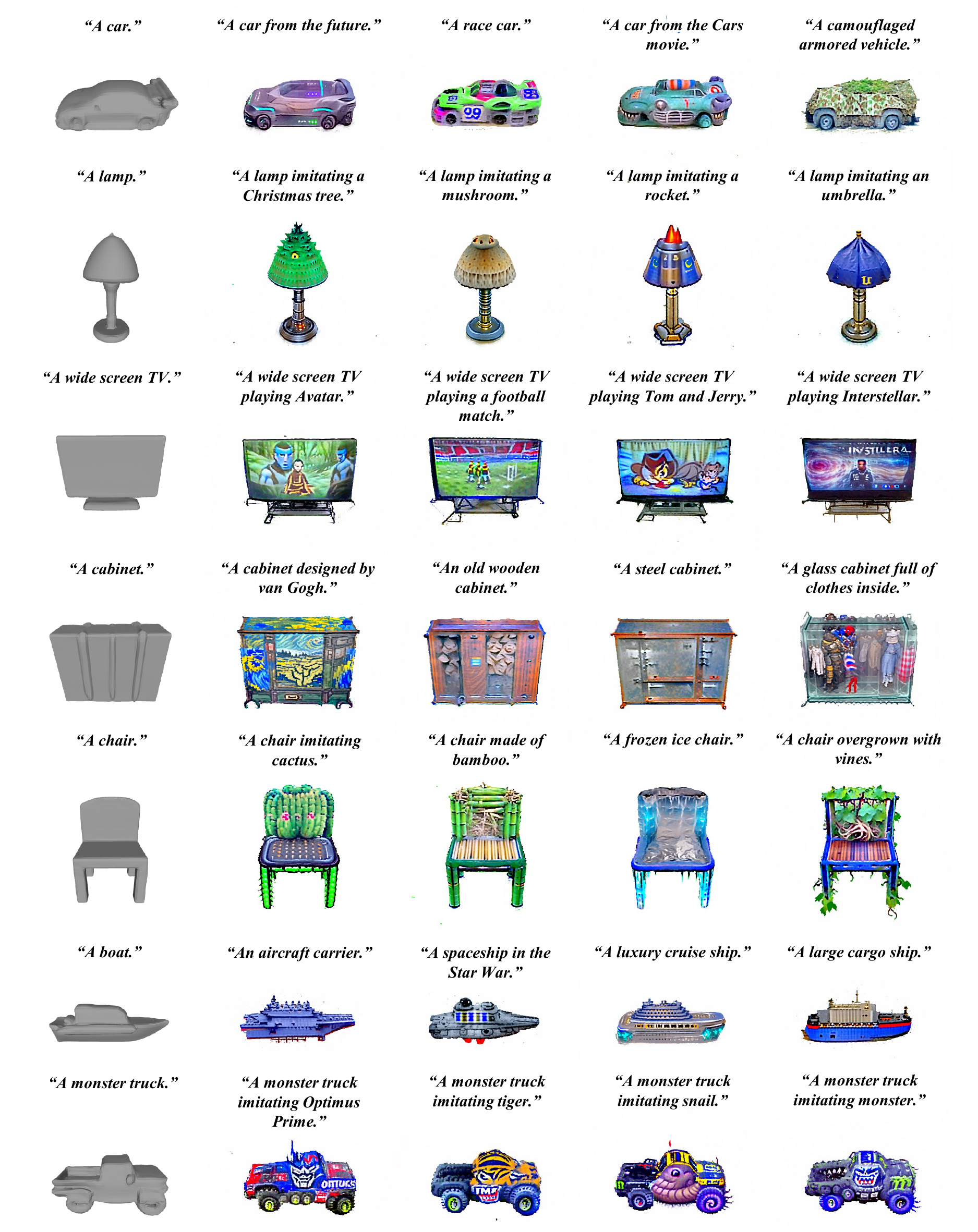}
   \caption{Additional text-to-3D synthesis results. We visualize the 3D shape prior used for optimization in the first column.}
   \label{fig_supp:text23d}
\end{figure*}

In this section, we show additional qualitative comparison results on text-to-3D synthesis with baseline methods in \cref{fig_supp:text23d_comparison} and more diversified generation results of our method in \cref{fig_supp:text23d}. It can be seen that our method can synthesize plausible 3D structures with the help of 3D shape priors. To better visualize the 3D structures generated by different methods, we also show video examples in the attached MP4 file.

% \section{3D Optimization with Inaccurate 3D Shape Prior}
% \label{sec_supp:inaccurate}

\clearpage
\clearpage
\newpage

\section{Integration with SVR Models}
\label{sec_supp:svr}

\subsection{Text-to-Shape Generation using SVR models}

\begin{figure}[th]
  \centering
  % \vspace{-0.5cm}
   \includegraphics[width=\linewidth]{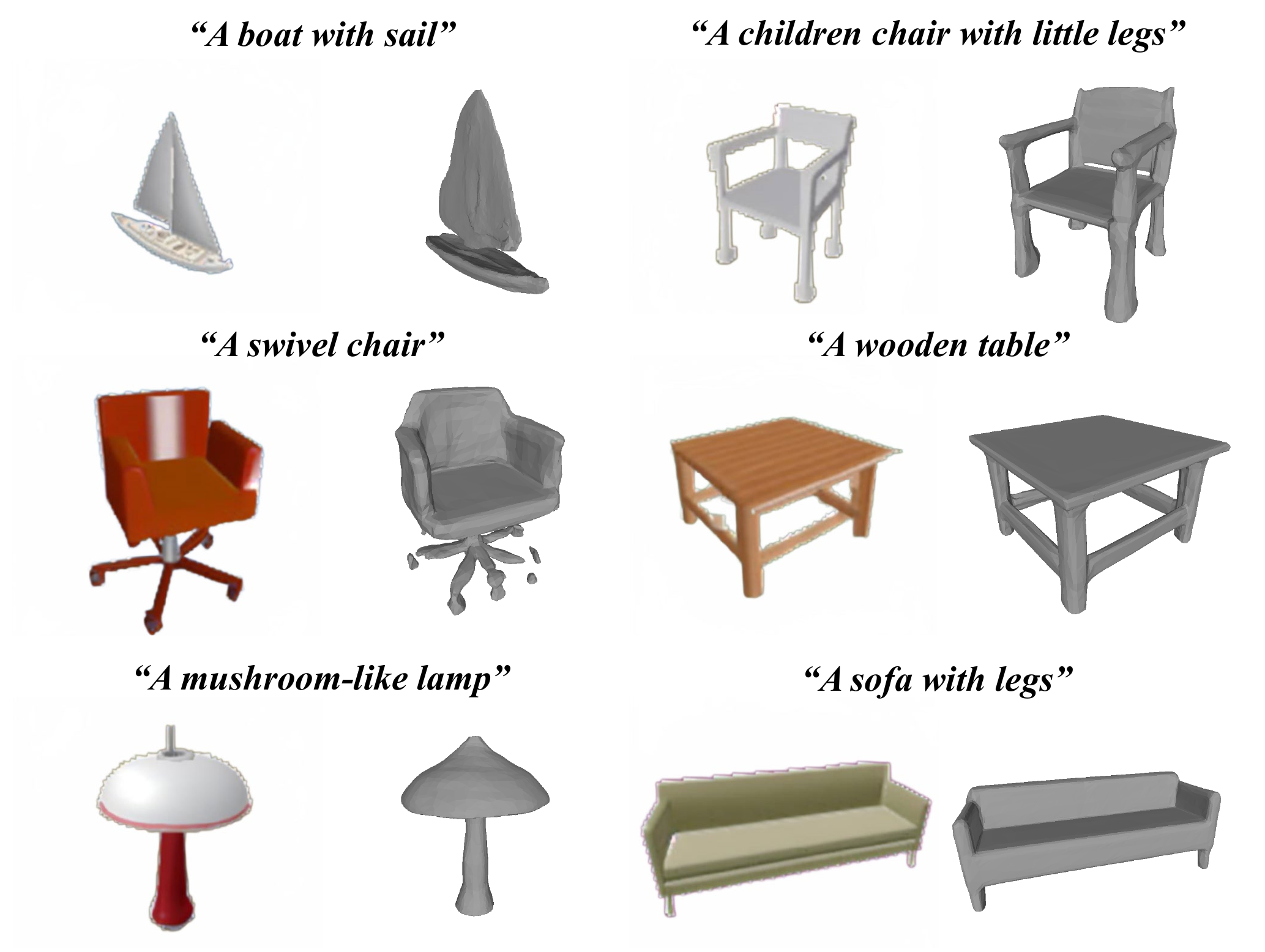}
   \caption{Text-guided shape generation using fine-tuned Stable Diffusion and Occupancy Networks~\cite{Mescheder_2019_CVPR}. For each text prompt, we visualize the image synthesized by the fine-tuned Stable Diffusion on the left and the reconstructed shape on the right.}
   \label{fig_supp:svr_occ}
\end{figure}

\begin{figure}[th]
  \centering
   \includegraphics[width=\linewidth]{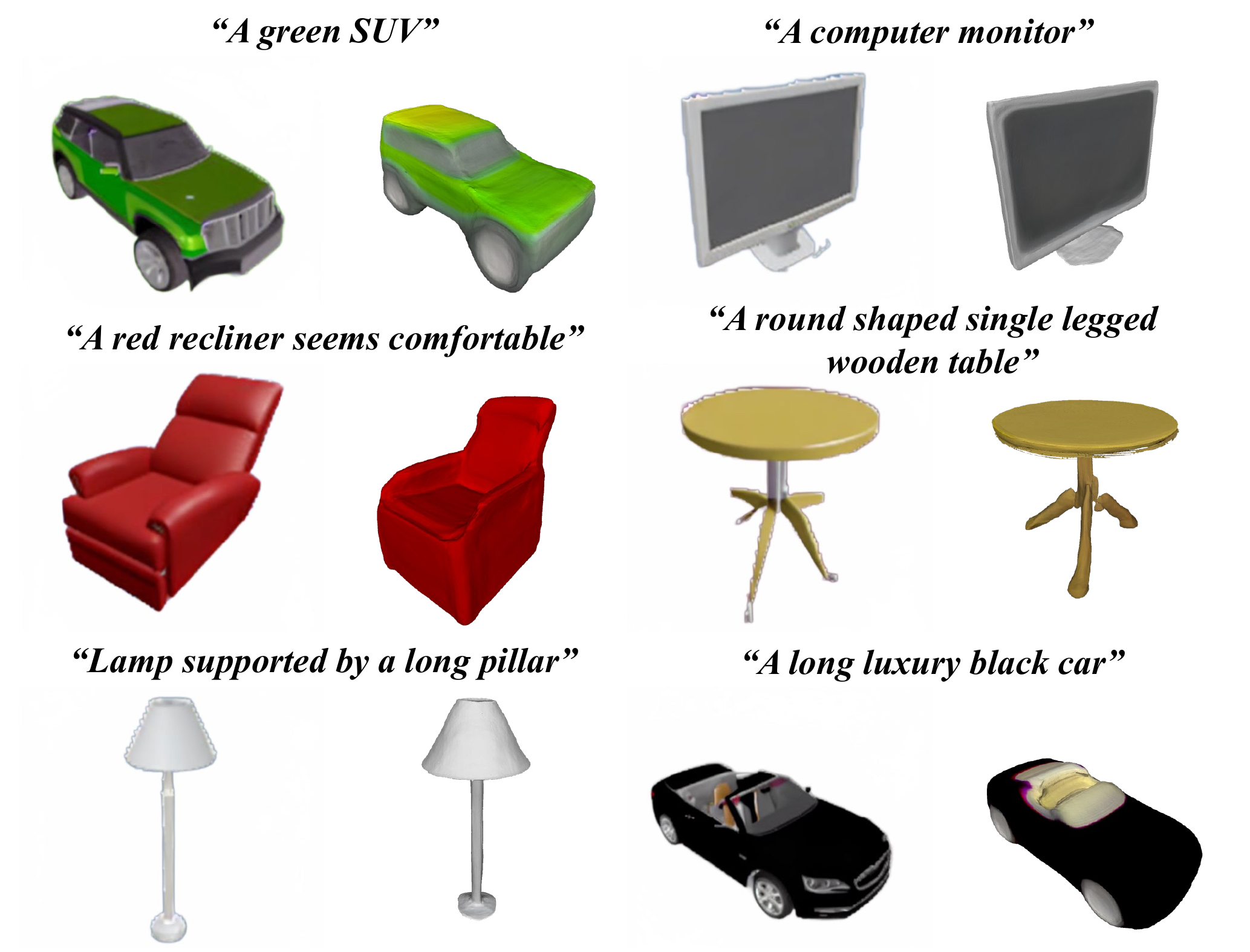}
   \caption{Text-guided shape generation using fine-tuned Stable Diffusion and DVR~\cite{Niemeyer_2020_CVPR}. For each text prompt, we visualize the image synthesized by the fine-tuned Stable Diffusion on the left and the reconstructed shape on the right.}
   \label{fig_supp:svr_dvr}
   % \vspace{-0.5cm}
\end{figure}

Single-view reconstruction (SVR) models can reconstruct a 3D shape from a single input image. We then ask, can we use an SVR model directly as the image-to-shape module in our framework? To answer this question, we conduct the same fine-tuning process to fine-tune a Stable Diffusion model with the ShapeNet renderings provided by Choy \etal~\cite{choy20163d} which are commonly used by many SVR methods. We find that although the shape renderings in Choy \etal~\cite{choy20163d} have more complex textures, the fine-tuned model can still capture the style successfully and synthesize novel images imitating the style. With such a fine-tuned Stable Diffusion, we can solve the text-to-shape generation in a precise way: synthesize an image using the fine-tuned Stable Diffusion with text prompt in the format of \textit{"a CLS in the style of *"}, and then directly feed the synthesized image into the SVR model. We show some text-guided shape generation results using two SVR methods, \ie, occupancy networks~\cite{Mescheder_2019_CVPR} and DVR~\cite{Niemeyer_2020_CVPR}, in \cref{fig_supp:svr_occ} and \cref{fig_supp:svr_dvr}, respectively. 
Both methods are trained with the shape renderings provided by Choy \etal~\cite{choy20163d}. 
The occupancy networks only predict shape, while DVR can predict both shape and color. As \cref{fig_supp:svr_occ} and \cref{fig_supp:svr_dvr} show, we achieve text-to-shape generation successfully with the synthesized images, which proves the strong generation ability of Stable Diffusion and the effectiveness of our fine-tuning pipeline.

A recent work named ISS~\cite{liu2022iss} also utilizes an SVR model to perform text-to-shape generation. However, the pipeline of ISS is much more complicated. It trains a mapper network to map CLIP features to the latent space of the SVR model, which requires a two-stage fine-tuning to align the text and shape feature spaces. At inference time, ISS needs to fine-tune the mapper network for each text prompt, which is redundant in our pipeline. With the help of the fine-tuned Stable Diffusion, we can directly generate an image from the text prompt and feed the image into the SVR model to synthesize a 3D shape. Besides, thanks to the strong generation ability of Stable Diffusion, we can enjoy a much larger generation diversity and synthesize as many 3D shapes as we want for each text prompt.

\begin{figure*}[th]
  \centering
  % \vspace{-0.5cm}
   \includegraphics[width=0.9\linewidth]{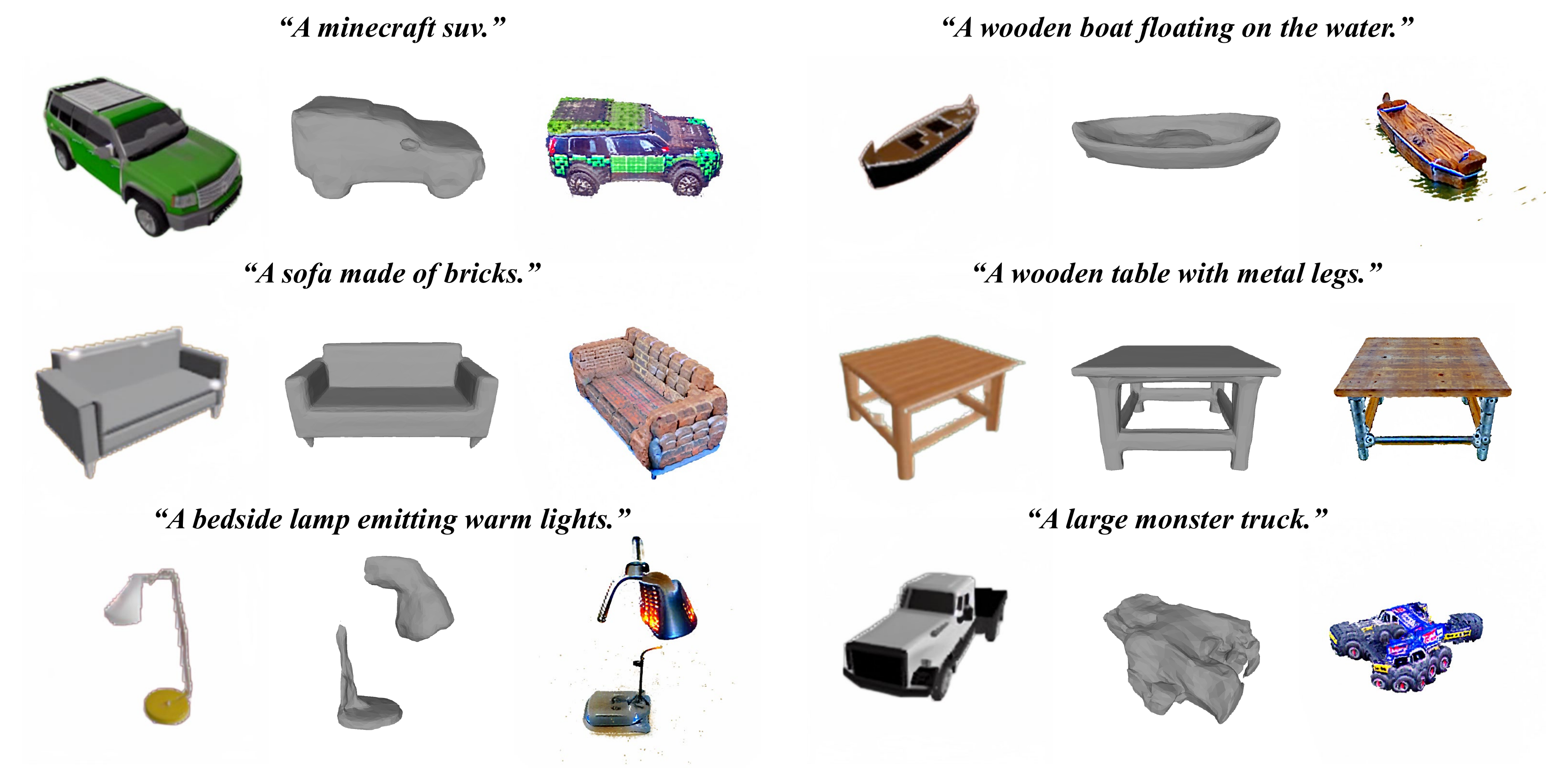}
   \caption{Text-to-3D synthesis results using occupancy networks. For each text prompt, we visualize the shape rendering image synthesized by the fine-tuned Stable Diffusion on the left, the shape reconstructed by the SVR model in the middle, and the optimization result on the right. The last row shows two failure cases.}
   % \vspace{-0.5cm}
   \label{fig_supp:svr23d}
\end{figure*}

\subsection{Text-to-3D Synthesis using SVR models}
Despite the success in text-guided shape generation with SVR models, we find that current SVR models are very sensitive to the input images. Although we can successfully capture the style of the shape renderings using the fine-tuned Stable Diffusion, some minor flaws in the synthesized images such as offsets of the objects from the image center and unrealistic artifacts (\eg, a chair lacks a leg) are inevitable. 
These minor flaws may lead to failed shape reconstructions, whose quality affects 3D shape priors. This sensitiveness makes the 3D prior generation in the first stage of our framework unstable. Therefore, we choose to use a 3D generator associated with a shape embedding mapping network to generate 3D shapes in the latent shape embedding space, instead of directly using an SVR model in our framework.

We visualize six text-to-3D synthesis results using 3D shape priors produced by the occupancy networks~\cite{Mescheder_2019_CVPR} in \cref{fig_supp:svr23d}. The successful results in the first two rows show the probability of integrating as SVR model into our framework. In the last row, we show two failure cases in which the SVR model fails to reconstruct plausible 3D shape priors to illustrate the drawbacks of using SVR models. We can observe that the discontinuity in the ``bedside lamp" shape leads to discontinuity in the final optimization result, while the failed truck shape results in total chaos.

\end{document}